\PassOptionsToPackage{unicode}{hyperref}
\PassOptionsToPackage{hyphens}{url}
\documentclass[preprint,1p]{elsarticle}
\usepackage{geometry}
\usepackage{xcolor}
\usepackage{amsmath,amssymb}
\usepackage{iftex}
\usepackage[unicode,hidelinks]{hyperref}
\usepackage{bookmark}
\usepackage{caption}
\usepackage{subcaption}
\usepackage{sectsty}
\sectionfont{\Large\bfseries}
\subsectionfont{\large\bfseries}
\usepackage{multirow}
\hypersetup{
  pdftitle  = {Graph Transformer-Based Flood Susceptibility Mapping},
  pdfauthor = {Vemula Sreenath; Filippo Gatti; Pierre Jehel}
}

\ifPDFTeX
  \usepackage[T1]{fontenc}
  \usepackage[utf8]{inputenc}
  \usepackage{textcomp} 
\else 
  \usepackage{unicode-math} 
  \defaultfontfeatures{Scale=MatchLowercase}
  \defaultfontfeatures[\rmfamily]{Ligatures=TeX,Scale=1}
\fi
\usepackage{lmodern}
\ifPDFTeX\else
\fi
\IfFileExists{upquote.sty}{\usepackage{upquote}}{}
\IfFileExists{microtype.sty}{
  \usepackage[]{microtype}
  \UseMicrotypeSet[protrusion]{basicmath} 
}{}
\makeatletter
\@ifundefined{KOMAClassName}{
  \IfFileExists{parskip.sty}{%
    \usepackage{parskip}
  }{
    \setlength{\parindent}{0pt}
    \setlength{\parskip}{6pt plus 2pt minus 1pt}}
}{
  \KOMAoptions{parskip=half}}
\makeatother
\usepackage{listings}

\usepackage{longtable,booktabs,array}
\usepackage{calc} 
\usepackage{etoolbox}
\makeatletter
\patchcmd\longtable{\par}{\if@noskipsec\mbox{}\fi\par}{}{}
\makeatother
\IfFileExists{footnotehyper.sty}{\usepackage{footnotehyper}}{\usepackage{footnote}}
\makesavenoteenv{longtable}
\usepackage{graphicx}
\makeatletter
\newsavebox\pandoc@box
\newcommand*\pandocbounded[1]{
  \sbox\pandoc@box{#1}%
  \Gscale@div\@tempa{\textheight}{\dimexpr\ht\pandoc@box+\dp\pandoc@box\relax}%
  \Gscale@div\@tempb{\linewidth}{\wd\pandoc@box}%
  \ifdim\@tempb\p@<\@tempa\p@\let\@tempa\@tempb\fi
  \ifdim\@tempa\p@<\p@\scalebox{\@tempa}{\usebox\pandoc@box}%
  \else\usebox{\pandoc@box}%
  \fi%
}
\def\fps@figure{htbp}
\makeatother
\usepackage{bookmark}
\IfFileExists{xurl.sty}{\usepackage{xurl}}{} 
\hypersetup{
  hidelinks,
  pdfcreator={LaTeX via pandoc}}

\date{}

\begin{document}

\begin{frontmatter}

\title{Breaking the Black Box:
Inherently Interpretable Physics-Constrained Machine Learning With Weighted Mixed-Effects for Imbalanced Seismic Data}

\author[1]{Vemula Sreenath\corref{cor1}}
\ead{vemula.sreenath@centralesupelec.fr}

\author[1]{Filippo Gatti}
\author[1]{Pierre Jehel}

\affiliation[1]{%
  organization={Universit\'e Paris-Saclay, CentraleSup\'elec, ENS Paris-Saclay, CNRS, LMPS -- Laboratoire de M\'ecanique Paris-Saclay},%
  city={Gif-sur-Yvette},%
  postcode={91190},%
  country={France}%
}

\cortext[cor1]{Corresponding author}

\begin{abstract}
Ground motion models (GMMs) are critical for seismic risk mitigation 
and infrastructure design. Machine learning (ML) is increasingly applied to GMM development 
due to expanding strong motion databases. However, existing ML-based GMMs operate as 
``black boxes,'' creating opacity that undermines confidence in engineering decisions. 
Moreover, seismic datasets exhibit severe imbalance, with scarce large-magnitude near-field 
records causing systematic underprediction of critical high-hazard ground motions. Despite 
these limitations, research addressing both interpretability and data imbalance remains limited. 
This study develops an inherently interpretable neural network employing independent additive 
pathways with novel HazBinLoss and concurvity regularization. HazBinLoss integrates physics-constrainted 
weighting with inverse bin count scaling to address underfitting in sparse, high-hazard regions. 
Concurvity regularization enforces pathway orthogonality, reducing inter-pathway correlation. 
The model achieves robust performance: mean squared error = 0.6235, mean absolute error = 0.6230, 
and coefficient of determination = 88.48\%. Pathway scaling corroborates established seismological 
behaviors. Weighted hierarchical Student-t mixed-effects analysis demonstrates unbiased residuals 
with physically consistent variance partitioning: sigma components range from 0.26-0.38 (inter-event), 
0.12-0.41 (inter-region), 0.58-0.71 (intra-event), and 0.68-0.89 (total). The lower inter-event and 
higher intra-event components have implications for non-ergodic hazard analysis. Predictions exhibit 
strong agreement with NGA-West2 GMMs across diverse conditions. This interpretable framework advances 
GMMs, establishing a transparent, physics-consistent foundation for seismic hazard and risk assessment. 
\end{abstract}


\end{frontmatter}

\textbf{Plain Language Summary}

Ground motion models (GMMs) predict how strongly the ground will shake during an earthquake. 
They are essential for structural analysis, seismic design, and seismic risk assessment studies. 
Traditional machine learning (ML) approaches are popular to develop GMMs, due to expanding 
earthquake databases worldwide. However, they operate as ``black boxes,'' which are hard to interpret 
and trust, limiting their use in high-stake decisions. Additionally, these databases suffer from 
significant data imbalances: fewer large, critically damaging records near the fault compared to 
abundant, less severely damaging distant records. This study addresses both limitations by developing 
a transparent ML architecture using the HazBinLoss function and concurvity regularization. Each input 
factor (magnitude, distance, their interaction term, etc.) is processed separately and combined linearly, 
revealing exact contribution of each component. HazBinLoss assigns higher weights to critical near-field 
large magnitude records and lower weights to less-critical records, during training to prevent 
underprediction of the most damaging scenarios. Rigorous statistical analysis using weighted mixed-effects 
confirms unbiased predictions with physically consistent error distributions. The model captures established 
seismological principles and achieves performance comparable to established GMMs while maintaining transparency. 
This framework enables broader adoption of ML-based approaches for risk assessment studies and disaster planning. 
\noindent\rule{\linewidth}{0.4pt}\vspace{4pt}
\textbf{Highlights}
\begin{itemize}
\item
  Interpretable neural network using independent additive pathways that 
  reveal exact contribution of magnitude, distance, and site effects
\item
  HazBinLoss function with physics-constrained weighting prioritizes critical 
  high-magnitude near-field records to prevent hazard underprediction
\item
  Weighted Student-t mixed-effects analysis demonstrates physically 
  consistent uncertainty partitioning with implications for hazard analysis
\end{itemize}
\vspace{4pt}\noindent\rule{\linewidth}{0.4pt}

\section{1. Introduction}

Earthquakes are one of the most damaging natural disasters, accounting
for 25.6\% of global economic disaster losses since 1900 (United Nations
Office for Disaster Risk Reduction, 2025). This seismic risk is
mitigated through probabilistic seismic hazard analysis (PSHA), which
relies on ground motion models (GMMs) to develop predictive equations
for key intensity measures such as the spectral acceleration (PSA) and
their associated uncertainties (Baker et al., 2021). Traditionally, GMMs
are developed empirically by assuming a functional form based on
underlying seismological principles that incorporates source parameters
(e.g., moment magnitude M\textsubscript{w}, fault type), path effects
(e.g., distance terms), and site conditions (e.g., shear-wave velocity
V\textsubscript{S30}) to estimate these measures, including
single-station sigma and total sigma for aleatory uncertainty
quantification (Al Atik et al., 2010; Baker et al., 2021).

Prominent traditional GMMs, such as those by ASK14: Abrahamson et al.
(2014), BSSA14: Boore et al. (2014), CB14: Campbell and Bozorgnia
(2014), and CY14: Chiou and Youngs (2014), were developed using the
global shallow crustal NGA-West2 dataset. Lack of universal strict
guidelines for selecting the number of input parameters leads to
variability: simpler models like BSSA14 incorporate limited terms, such
as magnitude scaling and style of faulting for the source term,
regionalized geometric spreading and anelastic attenuation for path
effects, basin effects, and regionalized linear and nonlinear site
response. In contrast, more complex models like ASK14 additionally
consider hanging-wall effects, depth to top of rupture
(Z\textsubscript{TOR}), directivity effects (though not explicitly
modelled as independent parameters), and aftershock scaling, which can
significantly complicate the GMM and potentially increase overfitting
risks (Bindi, 2017). In summary, these traditional approaches employ
diverse parameterizations to predict intensity measures and quantify
associated uncertainties; however, fixed functional forms can limit
flexibility in capturing complex data patterns.

Addressing these limitations, with the large volume of records in recent
ground motion databases, alternative data-driven machine learning (ML)
approaches have emerged for GMM development (e.g., Derras et al., 2014,
2016; Dhanya and Raghukanth, 2018, Meenakshi et al., 2023; Mohammadi et
al., 2023; Sedaghati and Pezeshk, 2023; Zhu et al., 2023; Fayaz et al.,
2024; Somala et al., 2024; Sreenath et al., 2024; Alidadi and Pezeshk,
2025; Ding et al., 2025). As nonparametric methods, these approaches
impose fewer assumptions on the data, learn highly nonlinear patterns,
and perform well with larger datasets, potentially resulting in improved
generalization over traditional GMMs. ML approaches excel in
interpolation but perform poorly in extrapolating unseen magnitude and
distance values, particularly without additional constraints. One
effective method to ensure the targets satisfy appropriate scaling and
attenuation constraints on magnitude, depth, distance, and
V\textsubscript{s30} is to implement monotonic constraints, as
demonstrated in Sreenath et al. (2024) and Okazaki et al. (2021).
Despite these advantages, interpretability and transparency remain
primary obstacles for ML-based GMMs (Alidadi and Pezeshk, 2025),
limiting their adoption in risk assessment.

Bridging parametric and nonparametric GMMs, another class of hybrid GMMs
has been recently developed using symbolic learning (Chen et al., 2024;
Liu et al., 2025). Rather than predefining a functional form, these
models employ symbolic operations within a neural network for
data-driven equation discovery. The core mechanism uses symbolic
activation layers combined with L\textsubscript{0} regularization to
prune redundant weights and promote sparsity after fully connected
transformations, yielding inherently interpretable models with an
explicit equation as the final output (Chen et al., 2024; Liu et al.,
2025).

Miller (2019) defines interpretability as "\emph{the degree to which a
human can understand the cause of a decision}." Complementing Miller
(2019), Lipton (2018) describes transparency via simulatability (the
ability to contemplate the entire model), decomposability (breaking it
down into intuitive parts), and algorithmic transparency (visibility
into the learning process). Post-hoc explanations address these issues
by using surrogate methods to interpret developed models.
Garson\textquotesingle s (1991) global approach analyzes neural network
weights to compute relative input feature importance, as applied in
studies like Vemula et al. (2021) and Dhanya et al. (2018). Local
methods, which focus on individual predictions, include SHapley Additive
exPlanations (SHAP) (Lundberg and Lee, 2017), which assigns importance
scores via game theory by evaluating marginal feature contributions
across coalitions (e.g., Somala et al., 2024; Fayaz et al., 2024), and
Local Interpretable Model-Agnostic Explanations (LIME) (Ribeiro et al.,
2016), which approximates local behavior by fitting simple models to
perturbed samples (e.g., Gharagozlou et al., 2025). However, Lipton
(2018) cautions that such post-hoc interpretability may not fully reveal
model behavior on unseen real-world data and can potentially mislead.

ML-based models are often viewed as "black boxes," as opposed to being
interpretable. Even with strong metrics such as low mean squared error
(MSE) or high coefficient of determination (R\textsuperscript{2}), this
opacity hinders adoption, as it offers an incomplete view of model
behavior (Molnar, 2022). Rudin (2019) argues that ML models are not
inherently opaque; rather, these issues arise from poor network
architecture choices, and thoughtful designs can yield inherently
interpretable models, which allow humans to understand without post-hoc
methods, with comparable performance. For example, using fully connected
layers worsens opacity through dense and entangled connections, which
obscure individual neuron contributions and impede modular decomposition
(Rudin, 2019). Linear and generalized linear models, decision trees,
generalized additive models (GAMs) --- which model the target as a sum
of smooth functions of individual predictors (Hastie and Tibshirani,
1986; Agarwal et al., 2020), and modular neural networks are some
inherently interpretable networks. XGBoost, random forest, and fully
connected feedforward neural networks are not inherently interpretable
due to the ensemble of trees or entanglement of weights, which obscures
the contribution of individual inputs. To date, the symbolic approach
represents a key development in inherently interpretable GMMs by
yielding explicit equations, while fully data-driven ML-based approaches
remain lacking in inherent interpretability.

The Gutenberg-Richter power-law implies fewer high-magnitude events,
while observational biases result in scarcer near-field records.
Conventionally, GMMs are trained by optimizing log-likelihood, often via
MSE loss under a Gaussian assumption, which leads to overfitting on
abundant small-magnitude, far-field data and underfitting on sparse,
large-magnitude near-field records. Si and Midorikawa (2000) and
Morikawa and Fujiwara (2013) considered weights based on distance,
obtained through trial and error, with higher weights assigned to
near-field records than far-field ones. This imbalance particularly
affects fully data-driven ML-based GMMs with limited seismological
constraints on the input scaling and attenuation patterns. To overcome
the underfitting problem due to imbalanced data in such cases, Kubo et
al. (2020) applied weighting to records based on peak ground
acceleration (PGA) values via trial and error. However, this resulted in
overfitting; consequently, they used a traditional GMM as a base model
and developed an ML model on the residuals (differences between targets
and base model predictions) without weighting. Kuehn (2023) recently
developed a robust Bayesian regression model with
Student\textquotesingle s t-distribution for residuals to accommodate
outliers and non-Gaussian variability. In the frequentist approach, no
systematic methods have been proposed to overcome data imbalance other
than trial-and-error weighting of records.

To bridge these gaps in systematic data handling and inherent
interpretability, this study introduces two key innovations.

\begin{enumerate}
\def\labelenumi{\alph{enumi}.}
\item
  novel HazBinLoss function that assigns a weight to each record by
  combining PGA-based hazard importance with inverse weighting according
  to the number of records in its corresponding magnitude--distance bin.
\item
  A linearly additive GAM-based ``glass-box'' GMM (contrasting
  ``black-box'') for PGA and PSA predictions is developed using global
  shallow crustal data in a frequentist approach. Our proposed approach
  ensures additivity and decomposability throughout the network
  architecture by avoiding interacting pathways in all layers, achieving
  an end-to-end fully interpretable, fully data-driven model with
  monotonic constraints.
\end{enumerate}

\section{2. Strong Motion Database}

The present study develops a GMM using the global shallow crustal data
from various strong motion databases for the 5\% damping to PGA, peak
ground velocity (PGV), and response spectrum (PSA). The NGA-West2
(Ancheta et al., 2014) contains data corresponding to regions from
Western North America (e.g., California), China, Mediterranean (Italy,
Greece, and Turkey), Japan, New Zealand, Taiwan, and other regions;
however, California, Japan, and Taiwan regions dominate the database
accounting for nearly 71\%, 9.1\%, and 9.3\%, respectively. To obtain a
better representation of remaining regions, data from corresponding
regional flatfiles for Turkey (Sandıkkaya et al., 2024), New Zealand
(Hutchinson et al., 2024), and Italy (Lanzano et al., 2022) are
considered instead of those from the NGA-West2 database. Table S1
confirms that all databases employ fundamentally compatible processing
procedures. The following criteria are applied to screen the data and
obtain the final data for developing the model:

\begin{enumerate}
\def\labelenumi{\arabic{enumi}.}
\item
  Consider events corresponding to shallow crustal tectonics with
  Z\textsubscript{TOR} $\le$ 20 km.
\item
  Remove records with no metadata for inputs (magnitude, distance,
  V\textsubscript{s30}, Z\textsubscript{TOR}) and targets.
\item
  For New Zealand data, consider strong motion data with channels
  corresponding to BN and HN.
\item
  Only consider records with rupture distance (R\textsubscript{rup}) $\le$
  300 km and M\textsubscript{w} \textgreater{} 3.
\item
  Remove records with sensor depth \textgreater{} 2m and the longest
  usable periods \textless{} 5 s.
\item
  Remove problematic records, e.g., late P-trigger, spectral quality,
  pulse type records.
\item
  Finally, consider events with at least 5 records for computing
  reliable mixed effects.
\end{enumerate}

The final selection contains 20,998 records from California, China,
Italy, Japan, New Zealand, Taiwan, and Turkey: 3900, 228, 1037, 1723,
3820, 1485, and 8805 records, respectively. A total of 962 earthquake
events recorded at 3970 stations are present in the data that was
screened. Fig. 1 and Table 1 illustrate the ranges of magnitude, rupture
distance, and other input features for the screened global data.
Magnitudes span from 3.3 to 7.9, with California, New Zealand, and
Turkey nearly covering the full range, though sparsely for
M\textsubscript{w} \textgreater{} 7. In contrast, China, Italy, Japan,
and Taiwan primarily include higher magnitudes. Rupture distances extend
from very close to the fault ($\approx$ 0.02 km) to far-field up to 300 km
across most regions, except Italy and Taiwan, which have limited
far-field records; notably, fewer than 10 records are very close to the
fault with R\textsubscript{rup} \textless{} 0.6 km, so these are not
shown in Fig. 1. All regions encompass soft to hard rock sites
(V\textsubscript{s30} from $\approx$ 50 m/s to 1749 m/s), with Japan featuring
very soft sites near 50 m/s. V\textsubscript{s30} values correspond to
both directly measured and estimated proxies. Z\textsubscript{TOR}
values range from surface ruptures (0 km) to a maximum of 20 km, though
Japan includes only shallow ruptures and Taiwan lacks deep ones.

\begin{longtable}[]{@{}
  >{\centering\arraybackslash}p{(\linewidth - 18\tabcolsep) * \real{0.1388}}
  >{\centering\arraybackslash}p{(\linewidth - 18\tabcolsep) * \real{0.0917}}
  >{\centering\arraybackslash}p{(\linewidth - 18\tabcolsep) * \real{0.0958}}
  >{\centering\arraybackslash}p{(\linewidth - 18\tabcolsep) * \real{0.0961}}
  >{\centering\arraybackslash}p{(\linewidth - 18\tabcolsep) * \real{0.0960}}
  >{\centering\arraybackslash}p{(\linewidth - 18\tabcolsep) * \real{0.0970}}
  >{\centering\arraybackslash}p{(\linewidth - 18\tabcolsep) * \real{0.0958}}
  >{\centering\arraybackslash}p{(\linewidth - 18\tabcolsep) * \real{0.0966}}
  >{\centering\arraybackslash}p{(\linewidth - 18\tabcolsep) * \real{0.0959}}
  >{\centering\arraybackslash}p{(\linewidth - 18\tabcolsep) * \real{0.0963}}@{}}
\caption{Table 1. Input parameter ranges for the considered global
crustal database}\tabularnewline
\toprule\noalign{}
\multirow{2}{=}{\centering\arraybackslash \begin{minipage}[b]{\linewidth}\centering
Region
\end{minipage}} &
\multirow{2}{=}{\centering\arraybackslash \begin{minipage}[b]{\linewidth}\centering
Region

Flag
\end{minipage}} &
\multicolumn{2}{>{\centering\arraybackslash}p{(\linewidth - 18\tabcolsep) * \real{0.1919} + 2\tabcolsep}}{%
\begin{minipage}[b]{\linewidth}\centering
M\textsubscript{w}
\end{minipage}} &
\multicolumn{2}{>{\centering\arraybackslash}p{(\linewidth - 18\tabcolsep) * \real{0.1930} + 2\tabcolsep}}{%
\begin{minipage}[b]{\linewidth}\centering
R\textsubscript{rup} (km)
\end{minipage}} &
\multicolumn{2}{>{\centering\arraybackslash}p{(\linewidth - 18\tabcolsep) * \real{0.1923} + 2\tabcolsep}}{%
\begin{minipage}[b]{\linewidth}\centering
V\textsubscript{s30} (m/s)
\end{minipage}} &
\multicolumn{2}{>{\centering\arraybackslash}p{(\linewidth - 18\tabcolsep) * \real{0.1922} + 2\tabcolsep}@{}}{%
\begin{minipage}[b]{\linewidth}\centering
Z\textsubscript{TOR} (km)
\end{minipage}} \\
& & \begin{minipage}[b]{\linewidth}\centering
Min
\end{minipage} & \begin{minipage}[b]{\linewidth}\centering
Max
\end{minipage} & \begin{minipage}[b]{\linewidth}\centering
Min
\end{minipage} & \begin{minipage}[b]{\linewidth}\centering
Max
\end{minipage} & \begin{minipage}[b]{\linewidth}\centering
Min
\end{minipage} & \begin{minipage}[b]{\linewidth}\centering
Max
\end{minipage} & \begin{minipage}[b]{\linewidth}\centering
Min
\end{minipage} & \begin{minipage}[b]{\linewidth}\centering
Max
\end{minipage} \\
\midrule\noalign{}
\endfirsthead
\toprule\noalign{}
\multirow{2}{=}{\centering\arraybackslash \begin{minipage}[b]{\linewidth}\centering
Region
\end{minipage}} &
\multirow{2}{=}{\centering\arraybackslash \begin{minipage}[b]{\linewidth}\centering
Region

Flag
\end{minipage}} &
\multicolumn{2}{>{\centering\arraybackslash}p{(\linewidth - 18\tabcolsep) * \real{0.1919} + 2\tabcolsep}}{%
\begin{minipage}[b]{\linewidth}\centering
M\textsubscript{w}
\end{minipage}} &
\multicolumn{2}{>{\centering\arraybackslash}p{(\linewidth - 18\tabcolsep) * \real{0.1930} + 2\tabcolsep}}{%
\begin{minipage}[b]{\linewidth}\centering
R\textsubscript{rup} (km)
\end{minipage}} &
\multicolumn{2}{>{\centering\arraybackslash}p{(\linewidth - 18\tabcolsep) * \real{0.1923} + 2\tabcolsep}}{%
\begin{minipage}[b]{\linewidth}\centering
V\textsubscript{s30} (m/s)
\end{minipage}} &
\multicolumn{2}{>{\centering\arraybackslash}p{(\linewidth - 18\tabcolsep) * \real{0.1922} + 2\tabcolsep}@{}}{%
\begin{minipage}[b]{\linewidth}\centering
Z\textsubscript{TOR} (km)
\end{minipage}} \\
& & \begin{minipage}[b]{\linewidth}\centering
Min
\end{minipage} & \begin{minipage}[b]{\linewidth}\centering
Max
\end{minipage} & \begin{minipage}[b]{\linewidth}\centering
Min
\end{minipage} & \begin{minipage}[b]{\linewidth}\centering
Max
\end{minipage} & \begin{minipage}[b]{\linewidth}\centering
Min
\end{minipage} & \begin{minipage}[b]{\linewidth}\centering
Max
\end{minipage} & \begin{minipage}[b]{\linewidth}\centering
Min
\end{minipage} & \begin{minipage}[b]{\linewidth}\centering
Max
\end{minipage} \\
\midrule\noalign{}
\endhead
\bottomrule\noalign{}
\endlastfoot
California & 1 & 3.34 & 7.28 & 2.5 & 300 & 116 & 1464 & 0 & 20 \\
China & 2 & 5.5 & 7.9 & 0.05 & 296.2 & 227 & 648 & 0 & 14.3 \\
Italy & 3 & 5.5 & 6.9 & 0.03 & 200 & 200 & 1286 & 0 & 18.5 \\
Japan & 4 & 6.6 & 6.9 & 0.02 & 300 & 49 & 1432 & 0.2 & 4 \\
New Zealand & 5 & 3.3 & 7.8 & 0.7 & 300 & 120 & 1500 & 0 & 20 \\
Taiwan & 6 & 5.9 & 7.6 & 0.6 & 172 & 124 & 1525 & 0 & 12 \\
Turkey & 7 & 3.5 & 7.8 & 0.1 & 300 & 131 & 1749 & 0 & 19.7 \\
\end{longtable}

\begin{figure}
\centering
\includegraphics[width=6.28819in,height=3.39722in]{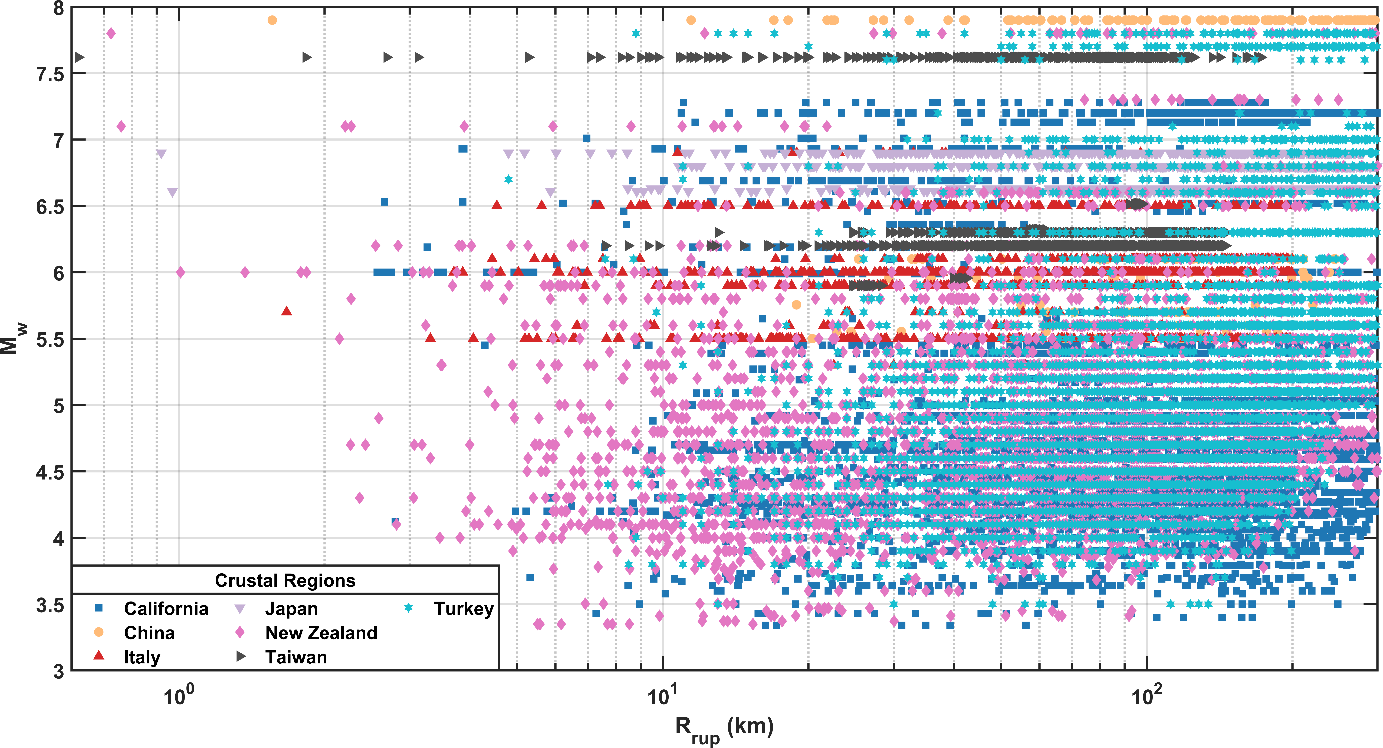}
\caption{Figure 1. Magnitude and rupture distance plot for different
regions in the screened data.}
\end{figure}

These uneven ranges across regions imply that predictions for some areas
may involve extrapolation (e.g., high magnitudes or deep
Z\textsubscript{TOR} in underrepresented regions), increasing
underfitting risks, while interpolation in well-covered areas could lead
to overfitting. Critically, the dataset has limited large-magnitude
near-field records compared to the abundance of lower-magnitude
far-field data, potentially causing the model to underfit in the most
damaging scenarios and overfit to the less damaging scenarios. These
data characteristics reveal inherent imbalances that pose challenges for
model training.

Fault classification is decided based on the rake angle ($\lambda$), with events
having -150 \textless{} $\lambda$ \textless{} -30 corresponding to normal (fault
flag = 0), 30 \textless{} $\lambda$ \textless{} 150 corresponding to reverse
(fault flag = 1), and other scenarios corresponding to strike-slip
(fault flag = 2) classification. Records that did not have sedimentary
basin depth values (Z\textsubscript{1}, depth to the shear-wave velocity
of 1.0 km/s) are computed using the equation provided by Chiou and
Youngs (2014). Additionally, California, China, Italy, Japan, New
Zealand, Taiwan, and Turkey are assigned region flags ranging from 1 to
7, respectively, as shown in Table 1.

Fig. S1 plots histograms of several key input variables. The magnitude
distribution is right-skewed as discussed previously, with a peak number
of events between 4.2 and 4.6. A considerable decline in available
events between 7 and 8 is evident. The distance histogram is
left-skewed, showing a sharp decline in record counts near the fault
compared to the far-field, with an order-of-magnitude difference (e.g.,
thousands at intermediate distances vs. hundreds below 10 km). The data
also include a slightly lower number of very-soft soil records with
V\textsubscript{s30} \textless{} 100 m/s and deep basin records with
Z\textsubscript{1} \textgreater{} 1000 m. Similarly, fewer deeper
rupture events (Z\textsubscript{TOR} \textgreater{} 15 km) are available
in comparison to surface rupture events. Critically, limited
large-magnitude events very near the rupture are present, as only a
handful of records yield negative M\textsubscript{w} $\times$
ln(R\textsubscript{rup}) values.

We model magnitude scaling, style of faulting, magnitude saturation,
anelastic attenuation, geometric spreading, depth scaling, basin depth
effects, linear site amplification, and regional effects. Hanging-wall
effects, directivity effects, and aftershock scaling are not modelled
owing to the paucity of input data. Nonlinear site amplification is not
modelled as one needs predicted PGA as input, which necessitates
developing another model. The outputs of the model are the PGA and PSA
at 25 periods (0.01, 0.02, 0.03, 0.04, 0.05, 0.075, 0.1, 0.12, 0.15,
0.17, 0.2, 0.25, 0.3, 0.4, 0.5, 0.6, 0.7, 0.8, 0.9, 1, 1.25, 1.5, 2, 3,
4, and 5 s). The following section discusses the development of the
novel HazBinLoss.

\section{3. Implementation of HazBinLoss}

In this section, details of the proposed loss are described. Kubo et al.
(2020) initially considered weighting records based on PGA to address
data imbalance, but their final model used a traditional parametric GMM
as a base. Building upon their work, this study develops a novel loss
termed HazBinLoss (Eq. 1). By applying weights at the mini-batch level,
HazBinLoss mitigates underfitting in sparse, large-magnitude near-field
regions, thereby enhancing model generalization. Note that a single
weight is assigned to each record, consistent across all periods.
HazBinLoss comprises two complementary terms: a global hazard-informed
component and an adaptive bin count component. The hazard-informed
component based on PGA values is computed once for the entire training
set, while the bin count component is computed adaptively for each
mini-batch based on the number of records in each bin. Here, a bin
refers to a 2D map defined by magnitude and distance. The
M\textsubscript{w} bins are divided into ranges: 3.0--4.0, 4.0--5.0,
5.0--6.0, 6.0--7.0, 7.0--7.2, 7.2--7.4, 7.4--7.6, 7.6--7.8, and
7.8--8.0. The R\textsubscript{rup} bins are grouped as 0--20 km, 20--50
km, 50--100 km, and 100--300 km. These components are combined using a
hyperparameter ($\alpha$), given in Eq. 2. Finally, a sigmoid scaling is
applied to obtain W\textsubscript{b(k)} to ensure the weights vary
smoothly using Eq. 3 and its parameters chosen to avoid saturation. The
combined weights, as visualized in Fig. 2, combine bin count and hazard
importance to prioritize high-damaging large-magnitude near-field
regions through higher values in the top-left areas, while
de-emphasizing abundant low-magnitude far-field bins, elsewhere.

\begin{longtable}[]{@{}
  >{\raggedright\arraybackslash}p{(\linewidth - 2\tabcolsep) * \real{0.9224}}
  >{\raggedright\arraybackslash}p{(\linewidth - 2\tabcolsep) * \real{0.0776}}@{}}
\toprule\noalign{}
\begin{minipage}[b]{\linewidth}\raggedright
\[\mathcal{L}_{HazBin} = \frac{1}{N}\sum_{k = 1}^{N}{W_{b(k)}{\ .\left( {\widehat{y}}_{k} - y_{k} \right)}^{2}} + \lambda\left\| \theta \right\|_{2}^{2}\]
\end{minipage} & \begin{minipage}[b]{\linewidth}\raggedleft
(1)
\end{minipage} \\
\midrule\noalign{}
\endhead
\bottomrule\noalign{}
\endlastfoot
\end{longtable}

In Eq. 1, N is the mini-batch size, W\textsubscript{b(k)} is the
combined weight (size = N $\times$ 1) assigned to the bin containing sample k,
b(k) denotes the bin index function that maps sample k to its
corresponding 2D bin based on M\textsubscript{w} and
R\textsubscript{rup} values, \(y_{k}\) and \({\widehat{y}}_{k}\) are
target and predicted values for sample k (size = N $\times$ 27, where 27
corresponds to PGA and the number of PSA periods), $\lambda$ is the
L\textsubscript{2} regularization coefficient, and $\theta$ represents the
model parameters. The underlying MSE operates on N $\times$ 27 tensors, with
the weight broadcast across columns.

\begin{longtable}[]{@{}
  >{\raggedright\arraybackslash}p{(\linewidth - 2\tabcolsep) * \real{0.9224}}
  >{\raggedright\arraybackslash}p{(\linewidth - 2\tabcolsep) * \real{0.0776}}@{}}
\toprule\noalign{}
\begin{minipage}[b]{\linewidth}\raggedright
\[W_{b(k)}^{raw} = (1 - \alpha)B_{k} + \alpha H_{k}\]
\end{minipage} & \begin{minipage}[b]{\linewidth}\raggedleft
(2)
\end{minipage} \\
\midrule\noalign{}
\endhead
\bottomrule\noalign{}
\endlastfoot
\end{longtable}

In Eq. 2, $\alpha$ is a tunable hyperparameter ranging from {[}0, 1{]}, with $\alpha$
values close to zero implying higher weighting to data distribution
based on bin counts (B\textsubscript{k}) and $\alpha$ values close to 1
implying higher weighting to hazard importance (H\textsubscript{k}).

\begin{longtable}[]{@{}
  >{\raggedright\arraybackslash}p{(\linewidth - 2\tabcolsep) * \real{0.9224}}
  >{\raggedright\arraybackslash}p{(\linewidth - 2\tabcolsep) * \real{0.0776}}@{}}
\toprule\noalign{}
\begin{minipage}[b]{\linewidth}\raggedright
\[W_{b(k)} = \frac{1}{1 + \exp\left( - 4\left( W_{b(k)}^{raw} - 0.5 \right) \right)}\]
\end{minipage} & \begin{minipage}[b]{\linewidth}\raggedleft
(3)
\end{minipage} \\
\midrule\noalign{}
\endhead
\bottomrule\noalign{}
\endlastfoot
\end{longtable}

\subsection{3.1 Hazard-Informed Component}

The hazard-informed component draws inspiration from physics-constrained
neural networks. Similar to Kubo et al. (2020), who weighted records
based on PGA values to emphasize potential hazard, this component
assigns higher values to high-magnitude near-field records, which are
critical for hazard assessment but underrepresented due to the power-law
distribution discussed in the introduction. Lower values are assigned to
low-magnitude far-field records. To compute this, we fit a simple GMM
for PGA using magnitude and distance terms with a functional form
similar to Idriss (2014), given by Eq. 4. This equation is evaluated at
bin mid values to obtain PGA-based importance values and are normalized
by its maximum value to obtain H\textsubscript{k}, ranging from (0,
1{]}. Intuitively, higher PGA values (e.g., M\textsubscript{w}
\textgreater{} 7 and R\textsubscript{rup} \textless{} 20 km) imply a
higher hazard-informed component, as they receive elevated weights as
demonstrated in Fig. 2, subplot (b).

\begin{longtable}[]{@{}
  >{\raggedright\arraybackslash}p{(\linewidth - 2\tabcolsep) * \real{0.9224}}
  >{\raggedright\arraybackslash}p{(\linewidth - 2\tabcolsep) * \real{0.0776}}@{}}
\toprule\noalign{}
\begin{minipage}[b]{\linewidth}\raggedright
\[\ln(PGA) = c_{0}\left( M_{w} - c_{1} \right) + c_{2}\left( M_{w} - c_{1} \right)^{2} + \left( c_{3} + c_{4}M_{w} \right)\ln\left( \sqrt{R_{rup}^{2} + c_{5}^{2}} \right) + c_{6}R_{rup}\]
\end{minipage} & \begin{minipage}[b]{\linewidth}\raggedleft
(4)
\end{minipage} \\
\midrule\noalign{}
\endhead
\bottomrule\noalign{}
\endlastfoot
\end{longtable}

In Eq. 4, c\textsubscript{0} to c\textsubscript{6} are the coefficients
determined by the nonlinear regression. c\textsubscript{0} = 0.8959,
c\textsubscript{1} = 3.814, c\textsubscript{2} = -0.17,
c\textsubscript{3} = 2.683, c\textsubscript{4} = -0.263,
c\textsubscript{5} = 5.046, and c\textsubscript{6} = -0.006435 (MSE:
0.7307, MAE: 0.6777, R\textsuperscript{2}: 0.8332). This equation
considers first-order and second-order magnitude scaling, geometric
attenuation, anelastic spreading, and magnitude-distance dependent
attenuation terms. To reduce complexity, this functional form is
region-agnostic and independent of V\textsubscript{s30},
Z\textsubscript{TOR}, and other effects, avoiding additional nested
loops per mini-batch.

\includegraphics[width=6.3in,height=2.55in]{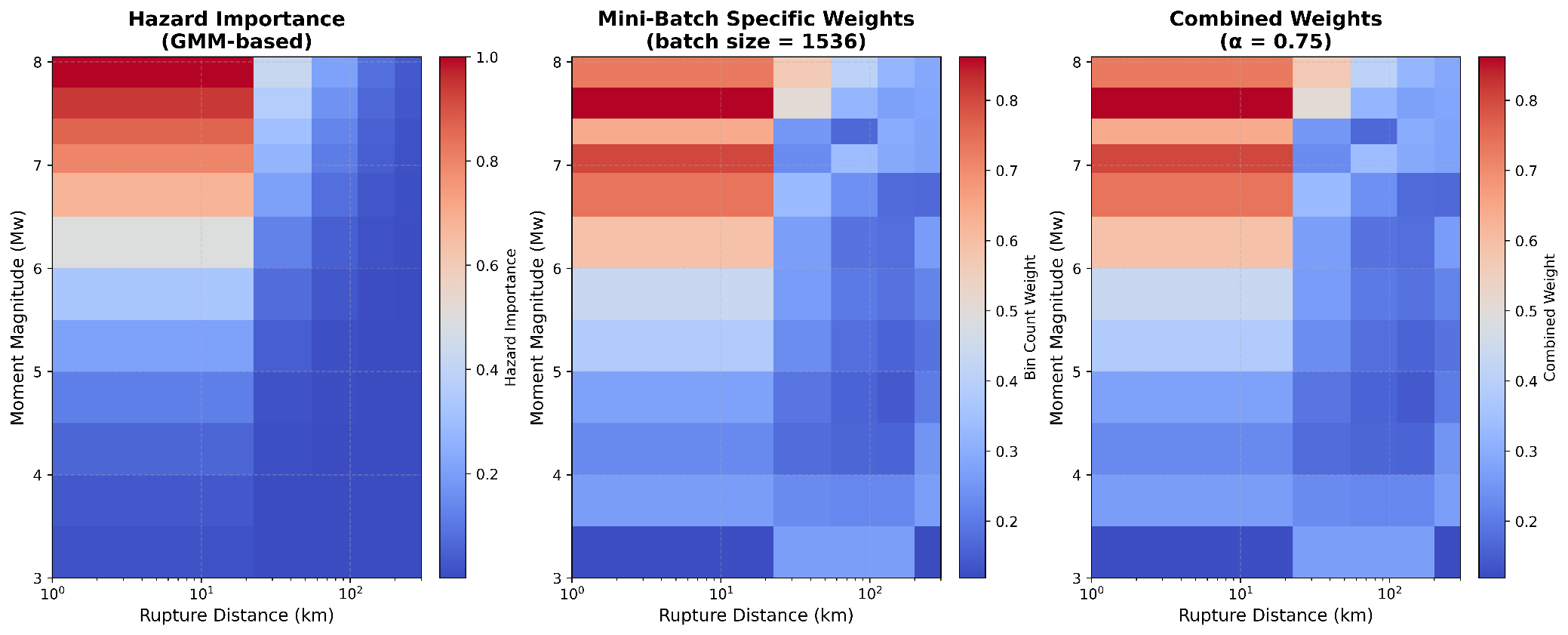}

Figure 2. The plot illustrates mini-batch weights, hazard importance
weights, and finally combined weights for a particular sample.

\subsection{3.2  Bin Count Component}

The bin count term addresses the data imbalance issue by weighing
inversely based on the sample density in each bin. It ensures that
abundant low-magnitude far-field records get lower weight, and sparse
high-magnitude near-field records get higher weight. For each bin
(\emph{i}, \emph{j}), the count weight C\emph{\textsubscript{i, j}} is
computed by Eq. 5. After trial and error, the logarithmic form (inspired
by inverse square root proportionality) is chosen as it smoothly scales
weights, with denser bins receiving lower values (but not too
aggressively low) to de-emphasize them. The values obtained from Eq. 5
are normalized by dividing by the maximum C\emph{\textsubscript{i, j}}
value to obtain B\textsubscript{k} values, which range from (0, 1{]}.
Fig. 2, subplot (a), illustrates this clearly for the bin count term,
conveying the above description.

\begin{longtable}[]{@{}
  >{\raggedright\arraybackslash}p{(\linewidth - 2\tabcolsep) * \real{0.9224}}
  >{\raggedright\arraybackslash}p{(\linewidth - 2\tabcolsep) * \real{0.0776}}@{}}
\toprule\noalign{}
\begin{minipage}[b]{\linewidth}\raggedright
\[C_{i,j} = 1 + \ln\left( \frac{\max\left( {count}_{k} \right)}{\max\left( {count}_{k},5 \right)} \right)\]
\end{minipage} & \begin{minipage}[b]{\linewidth}\raggedleft
(5)
\end{minipage} \\
\midrule\noalign{}
\endhead
\bottomrule\noalign{}
\endlastfoot
\end{longtable}

In Eq. 5, count\textsubscript{k} is the number of records in each 2D
magnitude-distance bin, max(count\textsubscript{k}) is the maximum count
across bins in the mini-batch (making it batch-adaptive), and a minimum
of 5 ensures near-empty bins are not excessively weighted. In the
following section, details of the developed architecture are discussed.

\includegraphics[width=6.29792in,height=5.28542in]{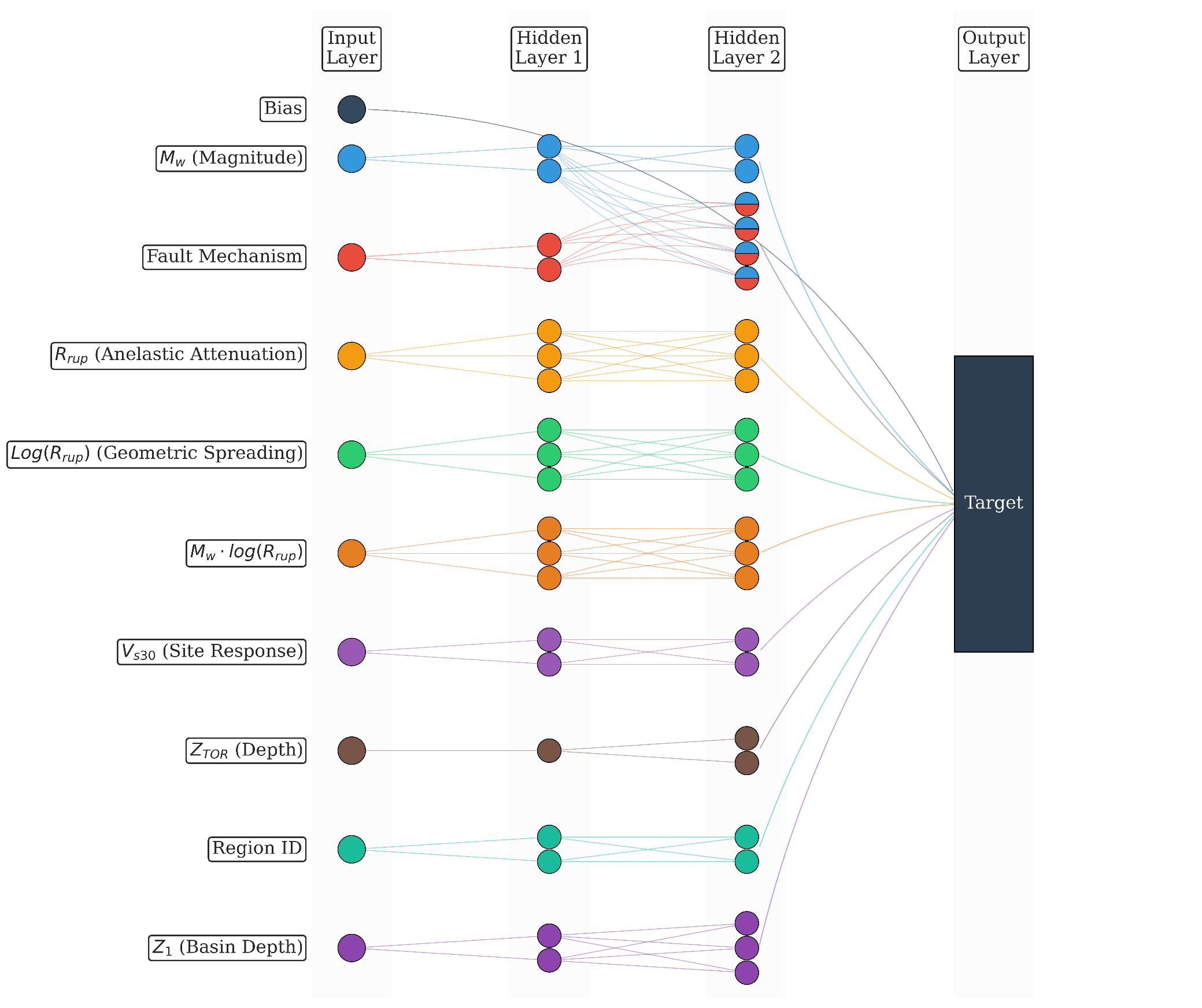}

Figure 3. Interpretable network architecture with independent pathways
for each input.

\section{4. Details of Interpretable Architecture}

In this section, we implement an inherently interpretable fully
data-driven neural architecture. Traditional neural networks can learn
complex nonlinear patterns on seismic data because input features are
allowed to entangle and have unrestricted interactions. However, this
obscures individual feature contribution to the final output, leading to
challenges such as decomposability and simulatability, as defined by
Lipton (2018), compromising transparency and hindering adoption in
hazard analysis. We address these critical issues by developing a
GAM-inspired neural architecture that processes input
features~additively~through non-interacting pathways (Hastie and
Tibshirani, 1986; Agarwal et al., 2020), ensuring end-to-end
interpretability while simultaneously taking advantage of data-driven
flexibility. This design allows for linear additivity across all the
input features without the need for post-hoc explanations, aligning with
Rudin (2019). In simple terms, each pathway corresponds to a neural
network with a single input, processing features independently before
additive summation to form the final prediction, given by Eq. 6. This
network architecture is illustrated in Fig. 3.

\begin{longtable}[]{@{}
  >{\raggedright\arraybackslash}p{(\linewidth - 2\tabcolsep) * \real{0.9224}}
  >{\raggedright\arraybackslash}p{(\linewidth - 2\tabcolsep) * \real{0.0776}}@{}}
\toprule\noalign{}
\begin{minipage}[b]{\linewidth}\raggedright
\[
\begin{aligned}
\ln(\mathrm{Target}) ={} &\, b + f_{\text{mag}}(M_w) + f_{\text{mag--FM}}(M_w,FM) + f_{\text{geo}}\!\bigl(\ln R_{\text{rup}}\bigr) + f_{\text{anelas}}(R_{\text{rup}}) \\[3pt]
&\; + f_{\text{linsite}}\!\bigl(\ln V_{s30}\bigr)+ f_{\text{depth}}(Z_{\text{TOR}})+ f_{\text{basin}}\!\bigl(\ln Z_{1}\bigr) \\[3pt]
&\; + f_{\text{mag--geo}}\!\bigl(M_w \,\ln R_{\text{rup}}\bigr)
+ f_{\text{reg}}(\text{Region flag}) + \mathrm{Residue}
\end{aligned}
\]
\end{minipage} & \begin{minipage}[b]{\linewidth}\raggedleft
(6)
\end{minipage} \\
\midrule\noalign{}
\endhead
\bottomrule\noalign{}
\endlastfoot
\end{longtable}

In Eq. 6, Target denotes the vector consisting of {[}PGA, PGV, PSA{]}, b
is a learnable bias, and each f( ) is a dedicated sub-network for each
effect we model and learned through backpropagation. Residue is the
remaining residual or irreducible noise corresponding to aleatory
variability, which cannot be explained with additional inputs or more
data. Additionally, monotonic constraints are enforced post-batch during
training to ensure physical consistency on magnitude, geometric and
anelastic distance, linear site, and depth layers. The implementation
details are provided in Sreenath et al. (2024) and Okazaki et al. (2021)
and thus are not provided here to avoid repetition. Therefore, the
proposed architecture captures: magnitude scaling using
M\textsubscript{w} term, style of faulting effects through
M\textsubscript{w} and FM flag terms, magnitude saturation through
M\textsubscript{w} $\times$ ln(R\textsubscript{rup}) term, anelastic
attenuation through R\textsubscript{rup} term, geometric spreading using
ln(R\textsubscript{rup}) term, depth scaling using Z\textsubscript{TOR},
basin depth effects using ln(Z\textsubscript{1}), linear site
amplification using ln(V\textsubscript{s30}), and regional effects using
regional flag term. It should be remembered that the first hidden layer
output from the f\textsubscript{mag}(M\textsubscript{w}) is passed as an
additional input to the second hidden layer of
f\textsubscript{mag-FM}(M\textsubscript{w}, FM) to enable controlled
interaction, while still maintaining interpretability. NGA-West2 models
consider FM dependent on M\textsubscript{w}. Hence, these two inputs are
considered for modeling the style of faulting term. To conclude, by
isolating effects into additive components, the network achieves
decomposability. The subsequent section discusses the performance of the
proposed architecture.

The earthquake events from the screened data are divided into 70\%,
15\%, and 15\% for training, validation, and testing sets, so that no
records from the same event would be split across sets, thereby
preventing intra-event correlation leakage and ensuring independent
evaluation. The model is trained in the PyTorch framework (Paszke et
al., 2019) with Optuna hyperparameter optimizer (Akiba et al., 2019).
Additionally, the work also makes use of the following Python libraries:
Pandas (McKinney, 2010), Numpy (Harris et al., 2020), and Matplotlib
(Hunter, 2007), and the R library: lme4 (Bates et al., 2015) for mixed
effects. Model training was conducted on an NVIDIA RTX A1000 6GB GPU
using CUDA 12.4, with each hyperparameter optimization run requiring
approximately 3 minutes, depending on learning rate and batch size.

\section{5. Results and discussion}

In this section, the performance of the proposed architecture and loss
function are evaluated. The network architecture (Fig. 3) employs two
hidden layers. The number of neurons in the pathways is treated as a
hyperparameter, ranging from 1 to 3. The second layer neurons count is
constrained to match the minimum neuron count from the first pathway. A
fewer neuron count is deliberately chosen to avoid overfitting without
sacrificing model performance.

Several network inputs are significantly correlated: M\textsubscript{w}
and M\textsubscript{w}-FM; R\textsubscript{rup} and
ln(R\textsubscript{rup}); V\textsubscript{s30} and Z\textsubscript{1};
M\textsubscript{w} and M\textsubscript{w}$\times$ln(R\textsubscript{rup}) to
name a few. Since the architecture employs a linear additive pathway, it
is essential to ensure that the variance is properly decomposed. It is a
critical issue for linear GMMs such as NGA models. However, despite
backpropagation, regularization, dropout ensure that the interpretable
GMM learns nonlinear mapping, variance decomposition is not guaranteed.
In this regard, concurvity analysis in R is performed. Concurvity is the
nonlinear analog of multicollinearity: strong correlations among
transformed features in additive models where one
pathway\textquotesingle s contribution can be approximated by
combinations of others, causing unstable decompositions (Siems et al.,
2023).

Fig. S2 displays period-averaged correlation coefficients between all
the output pathways for $\alpha$ = 0.25 model. It can be observed that as
discussed, pathway predictions of M\textsubscript{w} and
M\textsubscript{w}-FM, ln(V\textsubscript{s30}) and
ln(Z\textsubscript{1}), and several other combinations are significant
correlated, suggesting that the cannot reliably disentangle correlated
pathway contributions. Consequently, a~concurvity regularization term~is
introduced following Siems et al. (2023) to minimize variance leakage
between correlated predictors. The concurvity regularizer computes
the~mean absolute correlation coefficient~between all pairs of pathway
outputs, given by Eq. 7.

\begin{longtable}[]{@{}
  >{\raggedright\arraybackslash}p{(\linewidth - 2\tabcolsep) * \real{0.9224}}
  >{\raggedright\arraybackslash}p{(\linewidth - 2\tabcolsep) * \real{0.0776}}@{}}
\toprule\noalign{}
\begin{minipage}[b]{\linewidth}\raggedright
\[R_{\bot} = \frac{1}{n_{pairs}}\sum_{i < j}^{}\left| \rho\left( f_{i}(x),f_{j}(x) \right) \right|\]
\end{minipage} & \begin{minipage}[b]{\linewidth}\raggedleft
(7)
\end{minipage} \\
\midrule\noalign{}
\endhead
\bottomrule\noalign{}
\endlastfoot
\end{longtable}

In Equation 7, $\rho$ is correlation coefficient between pathways i and j,
f\textsubscript{i}(x) is output of pathway i, x is the input vector,
n\textsubscript{pairs} is the number of pathway pairs,
R\textsubscript{$\bot$} is penalized with regularization term
$\lambda$\textsubscript{concur}, a hyperparameter. Intuitively, minimizing
R\textsubscript{$\bot$} encourages the model to learn orthogonal independent
features for each pathway. However, the pathway independence is obtained
at the expense of slightly higher root mean-squared error (RMSE). For
instance, Siems et al. (2023) observed \textasciitilde10\% RMSE increase
on the California Housing dataset, while reducing pathway correlations,
R\textsubscript{$\bot$} from 0.20 to 0.05.

\begin{longtable}[]{@{}
  >{\raggedright\arraybackslash}p{(\linewidth - 2\tabcolsep) * \real{0.9224}}
  >{\raggedright\arraybackslash}p{(\linewidth - 2\tabcolsep) * \real{0.0776}}@{}}
\toprule\noalign{}
\begin{minipage}[b]{\linewidth}\raggedright
\[\mathcal{L}_{Total} = \mathcal{L}_{HazBin} + \lambda_{concur} \bullet R_{\bot}\]
\end{minipage} & \begin{minipage}[b]{\linewidth}\raggedleft
(8)
\end{minipage} \\
\midrule\noalign{}
\endhead
\bottomrule\noalign{}
\endlastfoot
\end{longtable}

\includegraphics[width=6.29792in,height=5.58958in]{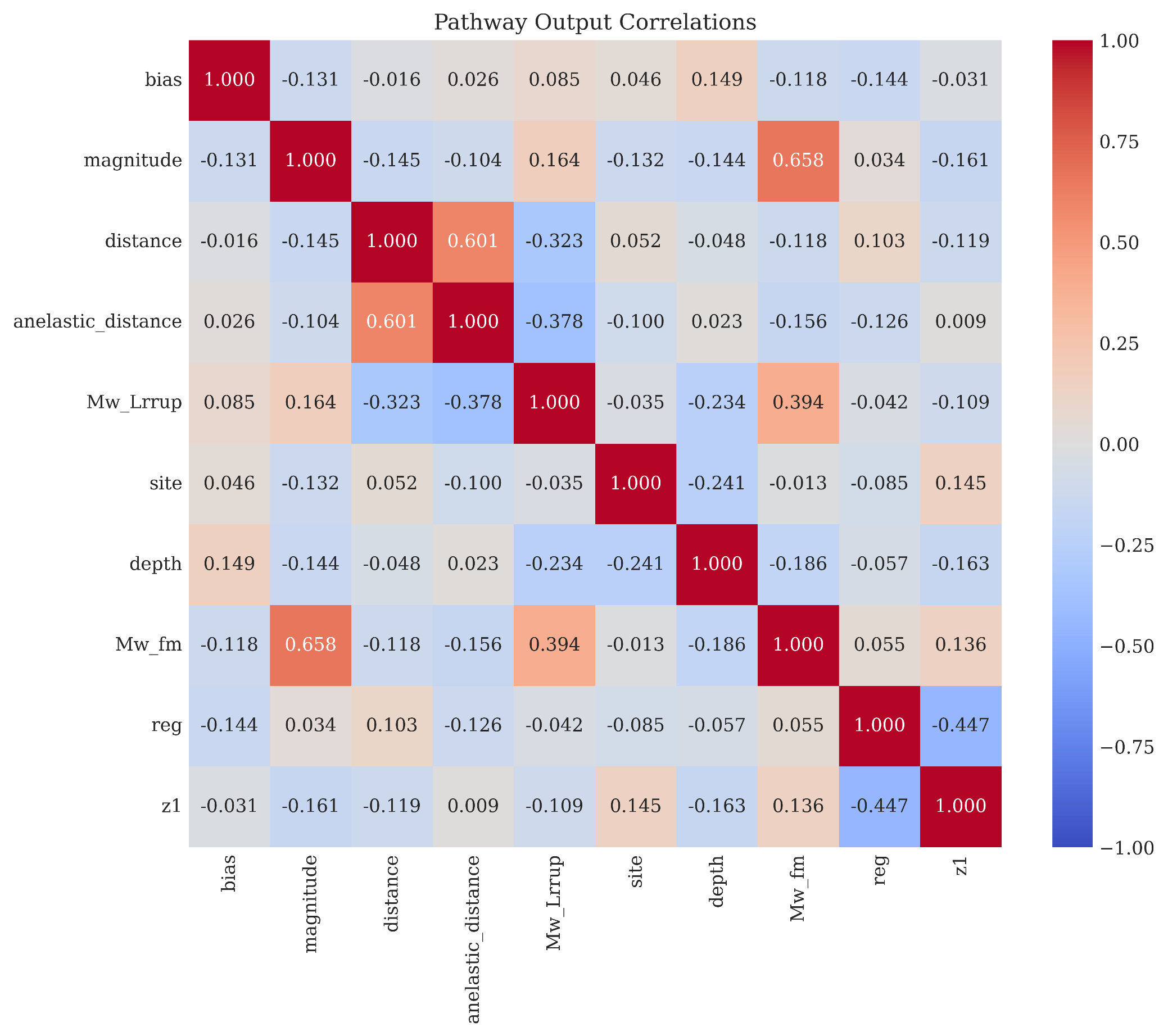}

Figure 4. Period-averaged correlation coefficients between all the
output pathways with concurvity regularization for $\alpha$ = 0.75 model.

The model is trained with Optuna hyperparameter tuning using the total
loss (Eq. 8) with $\alpha$ = 0, 0.25, 0.5, 0.75, and 1. The MSE, mean absolute
error (MAE), and R² metrics are then computed for each model using the
test records, as shown in Table 2. It can be observed that the MSE model
with concurvity loss achieves the best performance on the test records,
and the remaining models have comparable values. However, when the
metrics are computed using all the records corresponding to
M\textsubscript{w} $\ge$ 7 and R\textsubscript{rup} $\le$ 100 km, the MSE model
significantly underpredicts near-source ground motion records compared
to the models developed using any of the $\alpha$ values in HazBinLoss. It
implies that the models developed using the HazBinLoss need not result
in lower overall errors but do result in lower errors for
large-magnitude near-field records.

Table S2 compares the performance of MSE with concurvity loss model and
$\alpha$ = 0.75 model for different validation ranges, and we observe that $\alpha$ =
0.75 model consistently outperforms MSE model for all extreme bin
ranges. However, the performance gain becomes negligible with increasing
data considered for validation. For this problem, M\textsubscript{w} $\ge$ 7
and R\textsubscript{rup} $\le$ 100 km are selected based on statistical
robustness with hazard relevance. For a different database, this range
should be considered based on the data distribution. Furthermore,
robustness of bin definitions in HazBinLoss is assessed by training with
two alternative M\textsubscript{w}-- R\textsubscript{rup} partitions
(coarser and finer) and assessed model performance on M\textsubscript{w}
$\ge$ 7 and R\textsubscript{rup} $\le$ 100 km validation range and observed
consistent performance, indicating that the model performance is less
sensitive to bin definitions.

Fig. 4 displays the period-averaged correlation coefficients between all
the output pathways with concurvity regularization for $\alpha$ = 0.75 model.
It can be observed that the several pathways which were significantly
correlated are now moderately correlated with peak value of 0.961 (Fig.
S2) between M\textsubscript{w}-FM and M\textsubscript{w} is now 0.658.
Since, first layer magnitude and FM layer outputs are merged by design
resulting in some correlations after concurvity regularization. It is
thus recommended to consider M\textsubscript{w} $\times$ FM as an input instead
of combining output pathways, or increase network depth. It is
acknowledged that the interpretability is achieved by strictly additive
design but it cannot capture nonlinear cross-pathway interactions (e.g.,
soft-soil amplification that is PGA-dependent). This trade-off is
acceptable given our goal of transparent, physics-consistent
predictions. Alternative architectures (e.g., Neural Additive Models
with controlled interactions) could model nonadditive effects while
preserving interpretability---a direction for future work.

\begin{longtable}[]{@{}
>{\centering\arraybackslash}p{1.8cm}
>{\centering\arraybackslash}p{1.0cm}
>{\centering\arraybackslash}p{1.0cm}
>{\centering\arraybackslash}p{1.0cm}
>{\centering\arraybackslash}p{1.8cm}
>{\centering\arraybackslash}p{1.8cm}
>{\centering\arraybackslash}p{2.1cm}
>{\centering\arraybackslash}p{2.1cm}@{}}
\multicolumn{8}{@{}l@{}}{\textbf{Table 2.} Model performance using MSE with concurvity loss and total loss
for different $\alpha$ values} \tabularnewline[0.5em]
\toprule\noalign{}
\multirow{3}{=}{\centering Model\\(Loss)}
 & \multicolumn{3}{c}{\textbf{Test-set metrics}}
 & \multicolumn{2}{c}{$M_w\!\ge\!7$, $R_{rup}\!\le\!50$ km}
 & \multicolumn{2}{c}{High-mag bins} \\[2pt]
\cmidrule(lr){2-4} \cmidrule(lr){5-6} \cmidrule(lr){7-8}
 & MSE & MAE & $R^{2}$ & MSE & MAE
 & $M_w\!\ge\!7.8$, $R_{rup}\!\le\!20$ km
 & $M_w\!\ge\!7.8$, $R_{rup}\!\ge\!100$ km \\ 
\midrule
\endfirsthead
\toprule
\multicolumn{8}{c}{\small\emph{Table 2 --- continued from previous page}}\\
\midrule
\multirow{3}{=}{\centering Model\\(Loss)}
 & \multicolumn{3}{c}{\textbf{Test-set metrics}}
 & \multicolumn{2}{c}{$M_w\!\ge\!7$, $R_{rup}\!\le\!50$ km}
 & \multicolumn{2}{c}{High-mag bins} \\[2pt]
\cmidrule(lr){2-4} \cmidrule(lr){5-6} \cmidrule(lr){7-8}
 & MSE & MAE & $R^{2}$ & MSE & MAE
 & $M_w\!\ge\!7.8$, $R_{rup}\!\le\!20$ km
 & $M_w\!\ge\!7.8$, $R_{rup}\!\ge\!100$ km \\ 
\midrule
\endhead

\bottomrule\noalign{}
\endlastfoot
MSE Loss & 0.6021 & 0.6069 & 0.8737 & 0.4928 & 0.5566 & 1.1610 &
0.5980 \\
$\alpha$ = 0 & 0.6201 & 0.6205 & 0.8549 & 0.4511 & 0.5303 & 0.7920 & 0.5872 \\
$\alpha$ = 0.25 & 0.6113 & 0.6157 & 0.8720 & 0.4465 & 0.5270 & 1.1374 &
0.5865 \\
$\alpha$ = 0.5 & 0.6113 & 0.6264 & 0.8616 & 0.4291 & 0.5187 & 0.6908 &
0.6311 \\
\textbf{$\alpha$ = 0.75} & 0.6235 & 0.6230 & 0.8848 & \textbf{0.4286} &
\textbf{0.5183} & 0.6717 & 0.6137 \\
$\alpha$ = 1 & 0.6114 & 0.6217 & 0.8684 & 0.4395 & 0.5239 & 0.8954 & 0.5924 \\
\midrule
\addlinespace[3pt]
\multicolumn{8}{@{}>{\centering\arraybackslash}p{(\linewidth - 14\tabcolsep) * \real{1.0000} + 14\tabcolsep}@{}}{%
Models retrained, removing M\textsubscript{w} $\ge$ 6 and
R\textsubscript{rup} $\le$ 100 km data from training and validation
sets.} \\
\midrule
\addlinespace[3pt]
MSE Loss & 0.6366 & 0.6316 & 83.80 & 0.5138 & 0.5616 & 1.4007 &
0.7959 \\
$\alpha$ = 0 & 0.6621 & 0.643 & 83.74 & 0.5287 & 0.5677 & 1.1742 & 0.7618 \\
$\alpha$ = 0.25 & 0.6446 & 0.6374 & 85.98 & 0.4803 & 0.5461 & 1.0903 &
0.7934 \\
$\alpha$ = 0.5 & 0.6314 & 0.6313 & 86.08 & 0.4754 & 0.5414 & 1.1441 & 0.6858 \\
$\alpha$ = 0.75 & 0.6410 & 0.6595 & 84.16 & 0.4641 & 0.5383 & 0.9071 &
0.8021 \\
$\alpha$ = 1 & 0.6312 & 0.6522 & 84.93 & 0.4837 & 0.5499 & 0.9461 & 0.8833 \\
\end{longtable}

Furthermore, when evaluating the most critical records
(M\textsubscript{w} 7.8--8.0 and R\textsubscript{rup} 0--20 km), the MSE
loss model shows poor performance (MSE = 1.1610) compared to the optimal
HazBinLoss model with $\alpha$ = 0.75 (MSE = 0.6717). Additionally, since
near-field ground motions are dominated by source and path effects, and
nonlinear site effects are not considered, models with higher $\alpha$ values
achieve superior performance in these critical scenarios. In the far
field, all the models have similar performance metrics. We conclude that
the underweight assigned did not negatively influence the HazBinLoss
models to behave poorly at bins with ample records.

Ablation study: To further illustrate the efficacy of HazBinLoss, all
the records corresponding to M\textsubscript{w} $\ge$ 6 and
R\textsubscript{rup} $\le$ 100 km are removed from the training and
validation sets. The models are retrained without updating the Hazard
term in Eq. (8), and the results are shown in Table 2. The overall MSE
and MAE metrics, when computed for the test set for all the models, are
close, with a slight increase in error metrics compared to the previous
case due to data removal. Additionally, the MSE loss model performed
poorly, similar to the earlier case. Notably, the $\alpha$ = 0 model also
performs poorly because bin count weighting becomes ineffective without
actual high-magnitude near-field records to learn from. For other $\alpha$
values, the hazard importance term contributed to reducing the loss,
which demonstrates that the hazard-informed term enables learning about
critical scenarios even when direct training data is limited.

Model comparison:~A traditional data-driven feedforward network is
trained with HazBinLoss ($\alpha$ = 0.73), incorporating inputs:
M\textsubscript{w}, R\textsubscript{rup}, ln(R\textsubscript{rup}),
ln(V\textsubscript{s30}), ln(Z\textsubscript{1}), Z\textsubscript{TOR},
FM, and Region Flag (excluding the M\textsubscript{w} $\times$
ln(R\textsubscript{rup}) interaction term, as the weight interaction
captures this). The test set performance metrics for this model are: MSE
= 0.5612, MAE = 0.5954, and R² = 89.13\%. These results compare
favorably with the interpretable model ($\alpha$ = 0.75) metrics presented in
Table 2, with the fully-connected model achieving marginally superior
accuracy (\textasciitilde10\% increased error), due to concurvity
regularization and random weight initialization. For reference, the
performance metrics of established NGA-West2 models on the screened
dataset show: BSSA14 with MSE = 0.95 and MAE = 0.77, and CB14 with MSE =
0.88 and MAE = 0.73. Notably, the database contains a substantial
proportion of events from regions outside the original NGA-West2
dataset, which likely contributes to the elevated prediction errors
observed in these models.

It should be noted that since the model with $\alpha$ = 0.75 achieved better
performance on M\textsubscript{w} $\ge$ 7 and R\textsubscript{rup} $\le$ 100 km
data, it is considered for the subsequent analysis. The analysis
demonstrates that interpretable model architectures with independent
additive pathways achieve comparable predictive accuracy relative to the
fully data-driven approach and substantial performance improvements over
both NGA reference models.

\includegraphics[width=5.91025in,height=5.35119in]{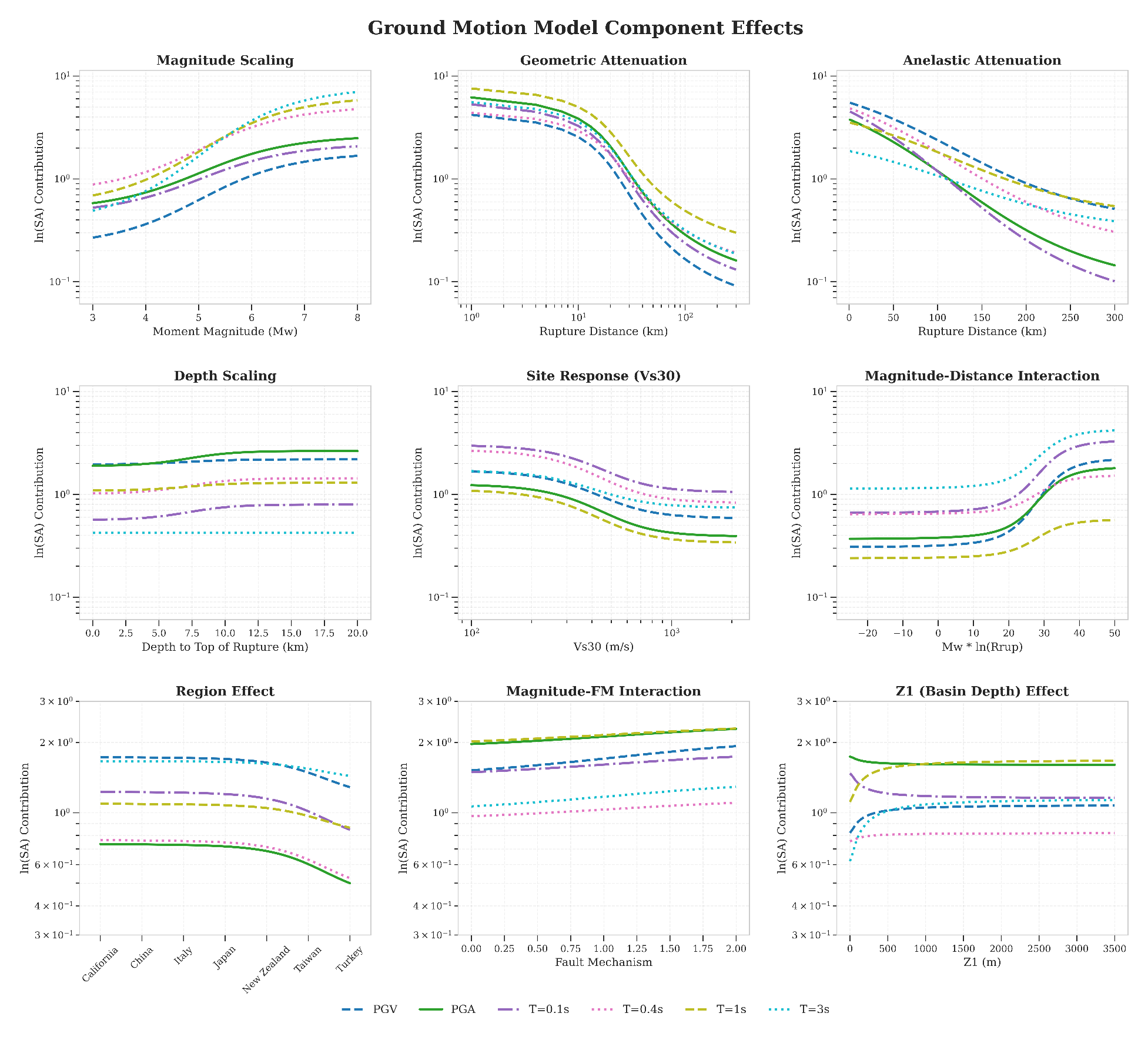}

Figure 5. Pathway scaling for the interpretable model with $\alpha$ = 0.75

\subsection{5.1 Interpretable Model Scaling}

This section analyzes the pathway scaling of the model with $\alpha$ = 0.75, as
illustrated in Fig. 5. All pathways exhibit period-dependent behavior
consistent with established seismological principles.

Source effects: The magnitude pathway term initially grows rapidly for
smaller magnitudes before saturating (for near-field distances) or
transitioning to linear growth (far-field distances), mirroring the
magnitude scaling in BSSA14, which exhibits parabolic growth up to a
hinge magnitude followed by saturation or linear behavior. The
magnitude-fault mechanism interaction exhibits linear scaling, with
strike-slip faulting producing the highest values, followed by reverse
and normal faulting, a pattern consistent with CB14. The depth scaling
term (Z\textsubscript{TOR}) increases up to $\approx$10 km before saturating at
lower periods, while remaining relatively constant at intermediate and
higher periods, differing from NGA-West2 models, which assumed linear
dependence across all periods.

Path Effects: The anelastic attenuation term exhibits linear dependence
on distance, similar to the region-dependent attenuation in BSSA14.
Geometric attenuation shows approximately linear behavior between 10-100
km with near-fault saturation effects similar to
BSSA14\textquotesingle s functional form, and steeper attenuation beyond
100 km. While NGA-West2 models employ the $\sqrt{R\textsuperscript{2}+h\textsuperscript{2}}$ term (where R is the
distance parameter and h is a fictitious distance parameter) to capture
near-fault saturation, the ln(R\textsubscript{rup}) term inherently
captures this effect. Similar conclusions are observed from Fig. S3,
where distance attenuation for different magnitude combinations is
illustrated.

Magnitude-Distance Effects: The magnitude-distance interaction term
exhibits saturation at lower values (corresponding to large-magnitude
near-field scenarios where saturation is expected) and linear growth at
higher M\textsubscript{w} $\times$ ln(R\textsubscript{rup}) values, consistent
with BSSA14\textquotesingle s M\textsubscript{w} $\times$ ln($\sqrt{R\textsuperscript{2}+h\textsuperscript{2}}$)
formulation. The model also captures other well-known phenomena. Lower
periods at higher magnitudes and near-field distances show converged
predictions, as evident in Fig. S3. Additionally, lower periods show
pronounced attenuation with the increase in distance. Finally, the gap
between different magnitude levels reduces as distance increases, with a
pronounced effect at the lower periods compared to the higher periods.

Site effects: The site response term shows a nonlinear decrease from
very-soft to rock sites before saturating at hard-rock conditions.
Higher values for very-soft sites imply significant spectral
amplification compared to other site conditions. Unlike the BSSA14
linear dependence assumption, our model captures the expected nonlinear
site response, which is evident from its non-uniform slope.
Additionally, period-dependent site amplification is evident. Basin
depth effects demonstrate exponential growth at shallow depths, followed
by saturation, consistent with CY14\textquotesingle s functional form.
Higher periods show significant basin amplification with limited
amplification at intermediate periods and de-amplification effects at
lower periods, as evident by the slope of the curves.

Region effects: Turkey exhibits the most negative regional adjustment
factors, corresponding to the highest attenuation relative to other
regions. Taiwan shows slightly higher values than Turkey, while
California demonstrates the highest amplification. China and Italy
exhibit regional amplifications similar to those in California.
Additionally, regional effects are most pronounced for high-magnitude,
far-field scenarios (e.g., M\textsubscript{w}=7.5,
R\textsubscript{rup}=300km) and less significant for low-magnitude,
near-field conditions (e.g., M\textsubscript{w}=3.5,
R\textsubscript{rup}=10km), as illustrated in Fig. S4. Regional
dependency exhibits strong period-dependence, with short periods showing
the highest regional variations across all magnitude-distance
combinations, while long periods demonstrate minimal regional
differences with overlapping response spectra. The regional attenuation
aligns with Danciu et al. (2024), which also found that Italian strong
motions generally attenuate more rapidly than those in the rest of
Europe. The model has learned region-specific path effects that could be
further enhanced by coupling the regional flag with the path terms.

\includegraphics[width=6.3in,height=4.99444in]{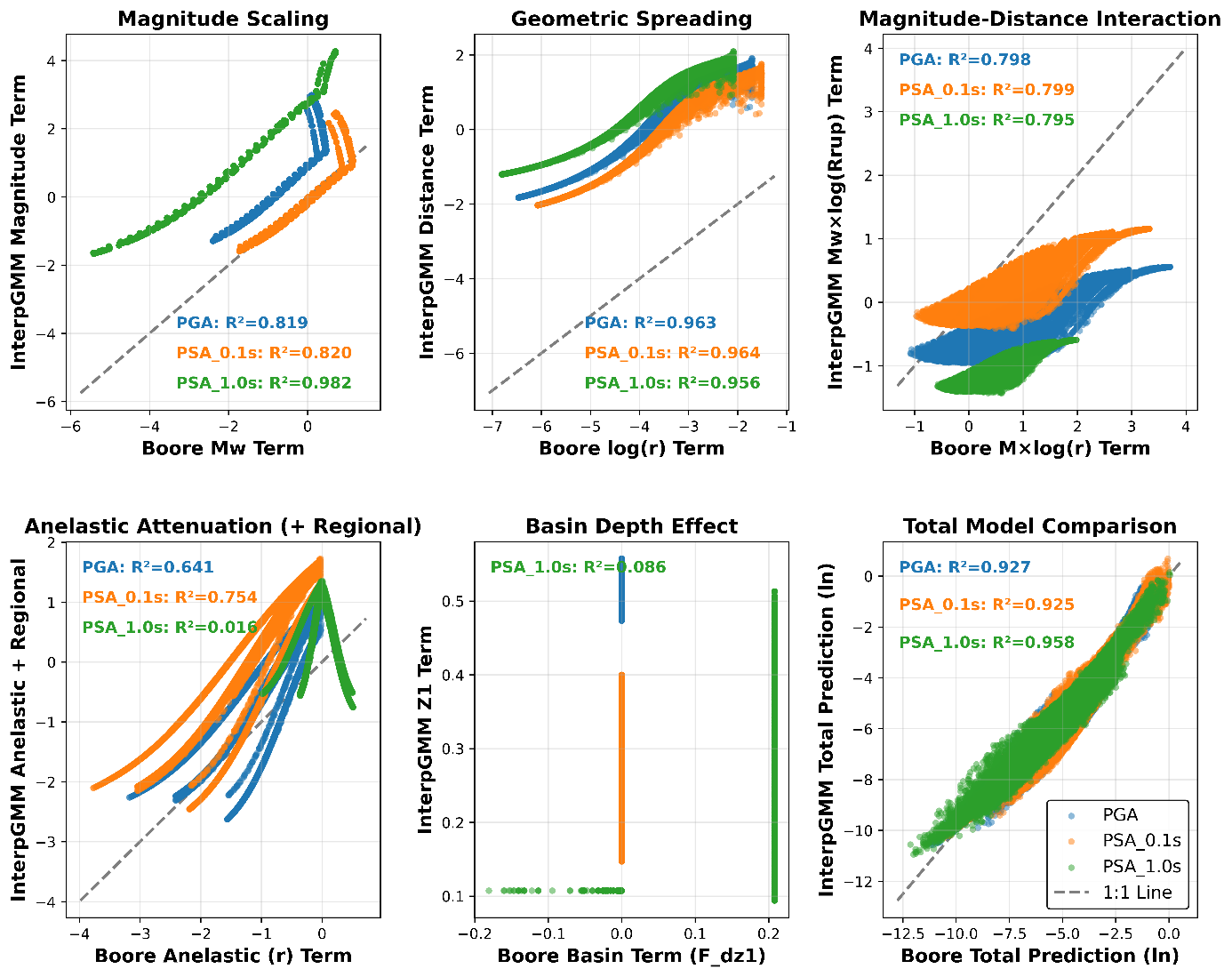}

Figure 6. Comparison of Boore et al. (2014) and interpretable GMM
pathways for entire database at PGA, PSA\textsubscript{0.1s}, and
PSA\textsubscript{1s}.

\subsection{5.2 Component Contribution: Interpretable model vs BSSA14}

In this subsection, the pathway component contributions from the
interpretable model are compared against BSSA14 model for PGA, PSA at
0.1s, and 1.0s in Fig. 6. Since BSSA14 employ a different functional
form (omit depth term, use Joyner-Boore distance, etc.), R² between the
pathway outputs from the interpretable model and the BSSA14 components
are used. It can be observed that the interpretable model exhibits
strong agreement (R² \textgreater{} 0.8 for most periods) with BSSA14
across periods and pathways. This validates that the decomposition is
physically meaningful, and not merely random variance splitting.

However, the near-zero R² for PSA\textsubscript{1s} is surprising, given
that both the models exhibit linear anelastic attenuation. Region-wise
analysis reveals: Global (R = 0.9946), California (0.9971), Japan
(0.9963), China/Turkey (-0.9930), Italy (0.9984). Since Turkey comprises
\textasciitilde42\% of records and shows perfect negative correlation,
the combined R² might have approached zero. Turkey regional-specific
model, CSI25: Ceken et al. (2025), is used to validate this regional
scaling accuracy. It is observed that R between interpretable and CSI25
model for anelastic pathway is around 0.9694. This confirms that
interpretable model correctly learned Turkey regional scaling and BSSA14
mischaracterizes Turkish attenuation, likely due to
NGA-West2\textquotesingle s limited regional coverage.

While SHAP despite having a sound mathematical basis, is at best an
approximation to the actual model. Unfortunately, can't be applied to
this model as it violates input independence assumption, so the SHAP
values would not be meaningful.

\includegraphics[width=6.29444in,height=4.16667in]{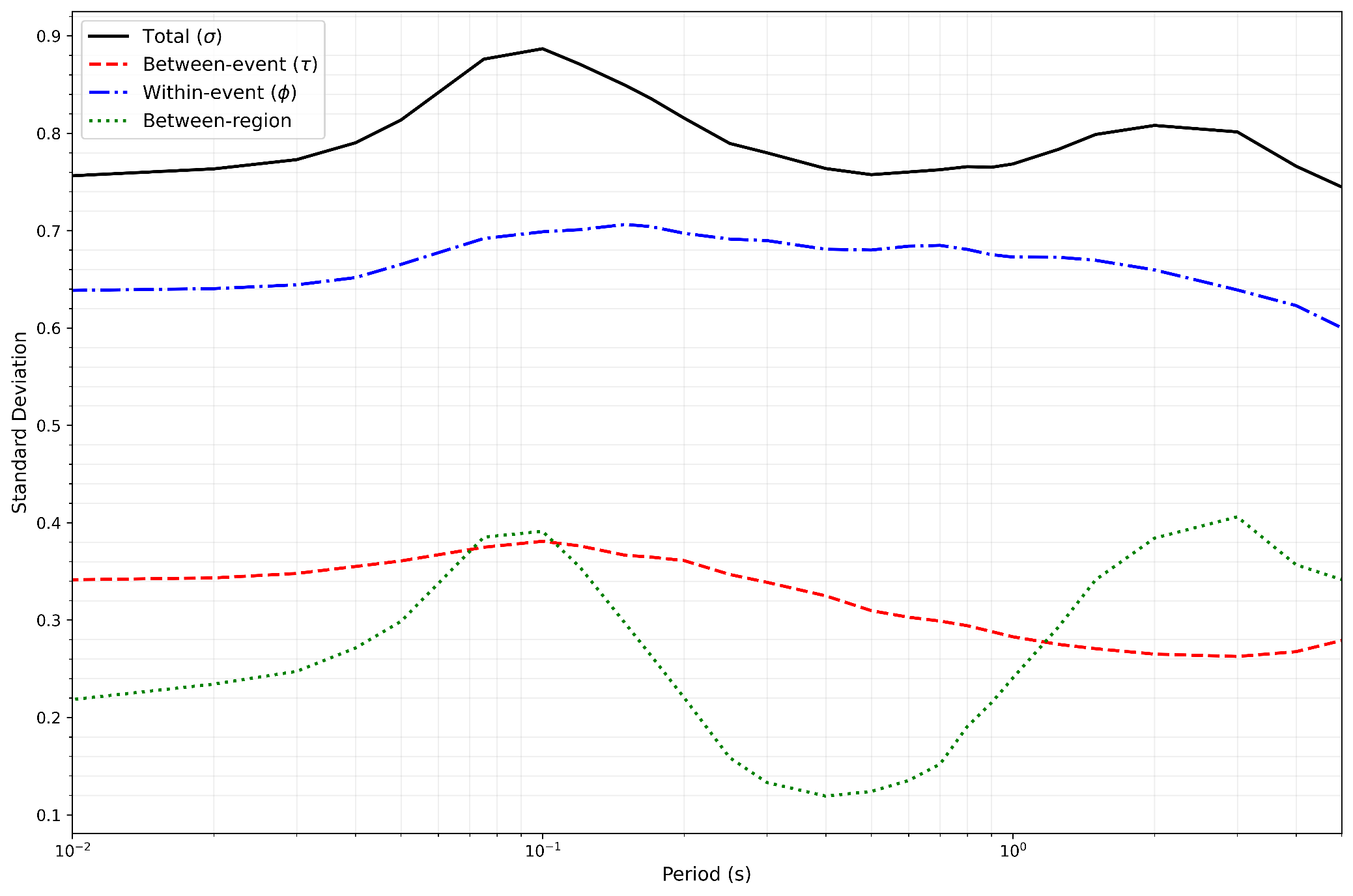}

Figure 7. Weighted mixed-effect regression analysis for the
interpretable model

\subsection{5.3 Mixed-Effect Analysis}

Earthquake records from the same seismic event recorded at multiple
stations exhibit similar properties due to common source
characteristics. Additionally, recorded data originating from the same
region and station have similar characteristics due to shared path and
site effects. The likelihood calculation in fixed-effects models ignores
this correlation structure, treating all records as independent
observations. This assumption results in significant bias in variance
estimates and predictions, as these records are not statistically
independent. Mixed-effects regression addresses this problem by
explicitly modeling the correlation structure, producing unbiased
residuals and predictions. Implementing mixed-effects for earthquake
data requires nonlinear mixed-effects modeling, which must be obtained
iteratively following the Abrahamson and Youngs (1992) approach. The
iterative algorithm alternates between fixed-effects parameter
estimation using neural networks and random-effects variance component
estimation using maximum likelihood in R\textquotesingle s glmmTMB
package.

We considered only earthquake events with at least 5 recordings to
obtain reliable random effects estimates. Of the 3,970 stations in the
screened database, 1,450 stations had only one record, and merely 981
stations had at least 5 recordings. Consequently, station terms are
excluded from the random effects structure. A hierarchical mixed-effects
model where earthquake events are nested within regions is developed
iteratively as given below (Meenakshi et al., 2023):

Iterative Nonlinear Mixed-Effects Algorithm:

\begin{itemize}
\item
  Input: X, y\textsuperscript{0}, R, E, $\epsilon$, max\_iter
\item
  Output~: $\theta$\textsuperscript{*}, $\sigma$, $\tau$, $\phi$\textsubscript{r}, $\phi$
\item
  Initialise: t $\leftarrow$ 0, y\textsuperscript{(0)} $\leftarrow$ y\textsuperscript{0}
\item
  Repeat:

  \begin{itemize}
  \item
    $\theta$\textsuperscript{(t+1)} $\leftarrow$ Train Interpretable Model ($\theta$\textsuperscript{(t)}; X, y\textsuperscript{(t)}, HazBinLoss)
  \item
    residual\textsuperscript{(t+1)} $\leftarrow$ y\textsuperscript{0} - f\textsubscript{$\theta$}(t+1)(X)
  \item
    \{$\delta$BR, $\delta$BE, $\delta$WER\} $\leftarrow$ MixedEffects(residual\textsuperscript{(t+1)} \textasciitilde{} 1
    + (1\textbar R/E))
  \item
    y\textsuperscript{(t+1)} $\leftarrow$ y\textsuperscript{(0)} - $\delta$WER
  \item
    t $\leftarrow$ t + 1
  \end{itemize}
\item
  until \textbar $\ell$\textsuperscript{(t)} - $\ell$\textsuperscript{(t-1)}\textbar{} \textless{} $\epsilon$ OR t $\ge$ max\_iter
\end{itemize}

In the above algorithm, X and y\textsuperscript{0} are the input and
target vectors, and y\textsuperscript{(t)} is the modified target. R and
E are vectors describing region and event details, $\epsilon$ and max\_iter are
set to be 5\% and 5. $\delta$BR, $\delta$BE, and $\delta$WER are random effects estimates for
between-region, between-event, and within-region-event components,
respectively. Residual: $\delta$BR + $\delta$BE + $\delta$WER. $\sigma$, $\tau$, $\phi$\textsubscript{r,} and
$\phi$ correspond to total sigma, between-event, between-region, and
within-region-event, respectively. $\theta$\textsuperscript{*} corresponds to
final weights of the model, and $\ell$\textsuperscript{t}: Log-likelihood value used for
convergence assessment.

To ensure consistency with the HazBinLoss weights assigns weights to
each record, same weights must be used in the mixed effects training.
The weighted mixed-effects regression assumes a Student-t distribution
for residuals with standard deviations: $\tau$, $\phi$\textsubscript{r}, and $\phi$,
where the total aleatory variability is given by Eq. 9. The $\nu$ (degrees
of freedom) values were close to 12-15 for the lower periods, indicating
substantial tail heaviness relative to Gaussian assumptions. Conversely,
higher periods exhibited $\nu$ values greater than 25, suggesting tail
behavior approaching Gaussian limits. The obtained standard deviations
are plotted in Fig. 7. They show a slightly higher sigma values than
NGA-West2 models due to the significantly larger database used for
training. The $\tau$, $\phi$\textsubscript{r}, $\phi$, and $\sigma$ values range in Fig. 7
between 0.2626-0.3809, 0.1192-0.4060, 0.5834-0.7064, and 0.6820-0.8868,
respectively.

\begin{longtable}[]{@{}
  >{\raggedright\arraybackslash}p{(\linewidth - 2\tabcolsep) * \real{0.9224}}
  >{\raggedright\arraybackslash}p{(\linewidth - 2\tabcolsep) * \real{0.0776}}@{}}
\toprule\noalign{}
\begin{minipage}[b]{\linewidth}\raggedright
\[\sigma\  = \sqrt{\tau^{2} + \ \phi_{r}^{2} + \ \phi^{2}}\]
\end{minipage} & \begin{minipage}[b]{\linewidth}\raggedleft
(9)
\end{minipage} \\
\midrule\noalign{}
\endhead
\bottomrule\noalign{}
\endlastfoot
\end{longtable}

The computed inter-event, inter-region, and intra-region-event residuals
are plotted to understand whether any bias exists in the developed model
with respect to input variables. Often, an underfit or overfit model
exhibits some bias. An underfit model fails to capture underlying
relationships, leading to systematic residual patterns across predictor
ranges. Fig. 8 illustrates the inter-region and inter-event residuals
plotted against magnitude and region flag, respectively, for PGA
(representing low periods), PSA at 0.3s (representing intermediate
periods), and PSA at 3s (representing high periods). The error bars are
plotted with the mean and standard deviation obtained from the residuals
within the bins considered. The mean values of inter-region residuals
are plotted at zero with standard deviation equal to the inter-region
residual value corresponding to each regional flag. The varying
inter-region standard deviations reflect the heterogeneity - regions
with minor standard deviations indicate more homogeneous seismic
characteristics, while larger standard deviations suggest greater
geological complexity or diverse recording conditions within that
region. Japan is observed consistently to exhibit greater diversity
compared to the other regions. No bias is evident from the inter-event
plots as the mean values are close to zero, and no trend is apparent
with inter-event standard deviations varying slightly across magnitude
bins. M\textsubscript{w} \textgreater{} 7

Fig. 9 similarly illustrates intra-region-event residuals plotted against magnitude and distance, 
respectively, for PGA, PSA at 0.3s, and PSA at 3s, along with error bars. Instead of plotting all 
residuals, only 2000 residuals are randomly chosen and plotted for clear visualization. The mean 
values at all bins are very close to zero in all subplots, and the standard deviations are consistent 
across bins. It demonstrates that the mixed-effects modeling approach successfully removes systematic 
biases with respect to magnitude and distance.

A critical distinction between this approach and traditional mixed-effects regression is that 
HazBinLoss assigns higher weights to large-magnitude near-field records and lower weights to 
moderate-magnitude records. This weighting has important implications for the obtained residual 
variance partitioning. To illustrate this, Student-t mixed-effects regression is performed without 
and with HazBinLoss weights. Fig. S5 and Fig. 7 show standard deviations plotted against period for 
unweighted and weighted analyses, respectively.

Total aleatory variability remains comparable between both analyses, indicating overall predictive 
uncertainty is preserved. Inter-regional residuals are also similar, as weighting was not applied 
between regions. However, inter-event and intra deviations differ significantly. The increase in 
intra deviations compared to BSSA14 and other NGA-West2 models is physically expected: high-magnitude 
events at close distances exhibit higher aleatory scatter due to complex source rupture, directivity 
effects, and near-fault path heterogeneity.

Furthermore, inter-event residuals exhibit magnitude-dependent bias in Fig. S6, particularly at M\textsubscript{w} > 6.5 
where data are limited. This artificial bias arises from treating all records equally during mixed-effects 
analysis when they were weighted during training. When weighting is consistently applied during both training 
and mixed-effects, the bias vanishes (Fig. 7). This variance partitioning has important implications for 
seismic hazard analysis: lower inter-event variability means less uncertainty reduction when multiple 
events contribute to a logic tree, while higher intra variability implies greater uncertainty 
in single-station ground motion estimates.

\includegraphics[width=6.3in,height=4.18333in]{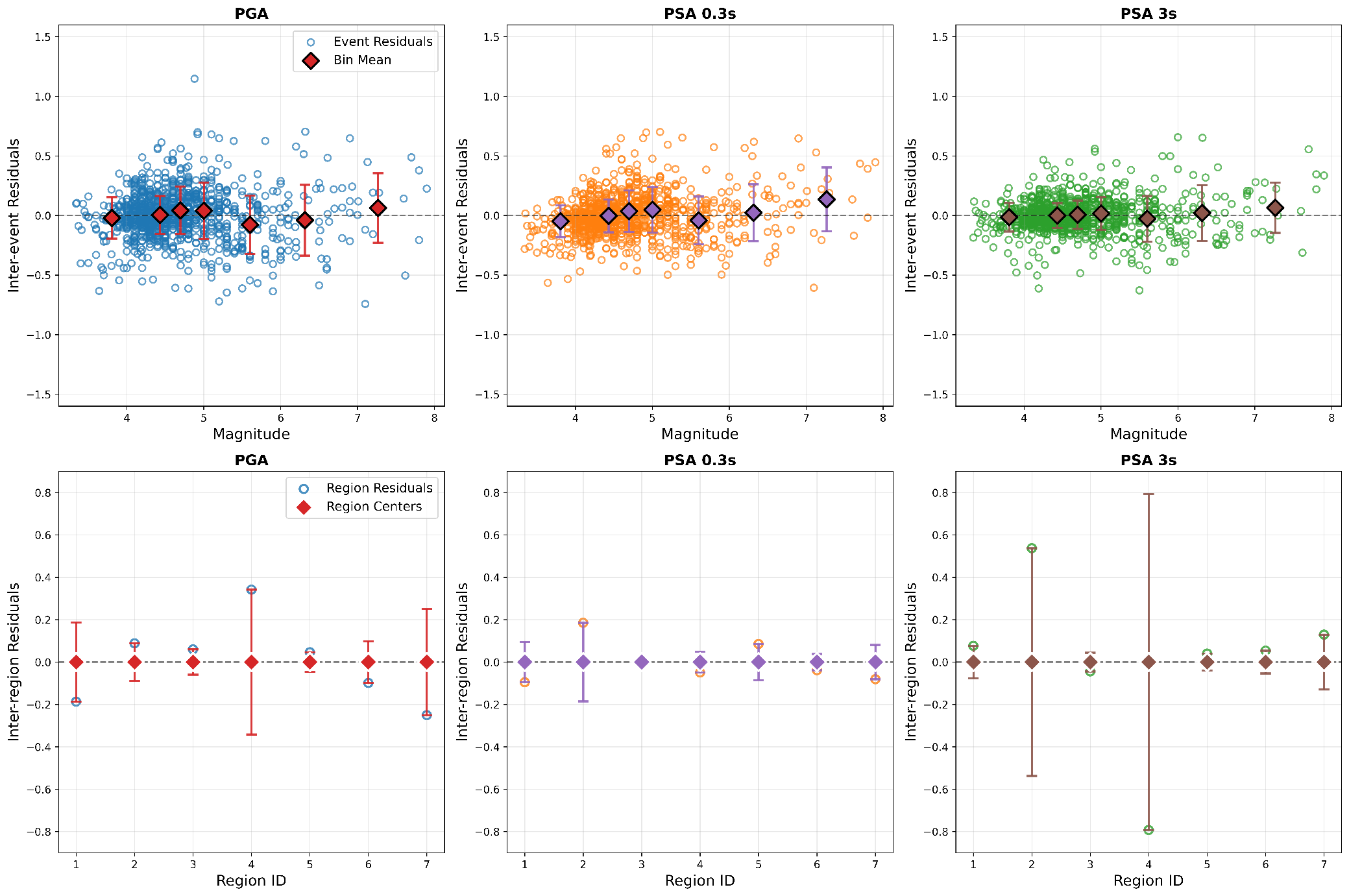}

Figure 8. Inter-event and inter-region residuals plotted against
magnitude and regional flags, respectively, for PGA, PSA at 0.3 and 3s.

\includegraphics[width=6.3in,height=4.18333in]{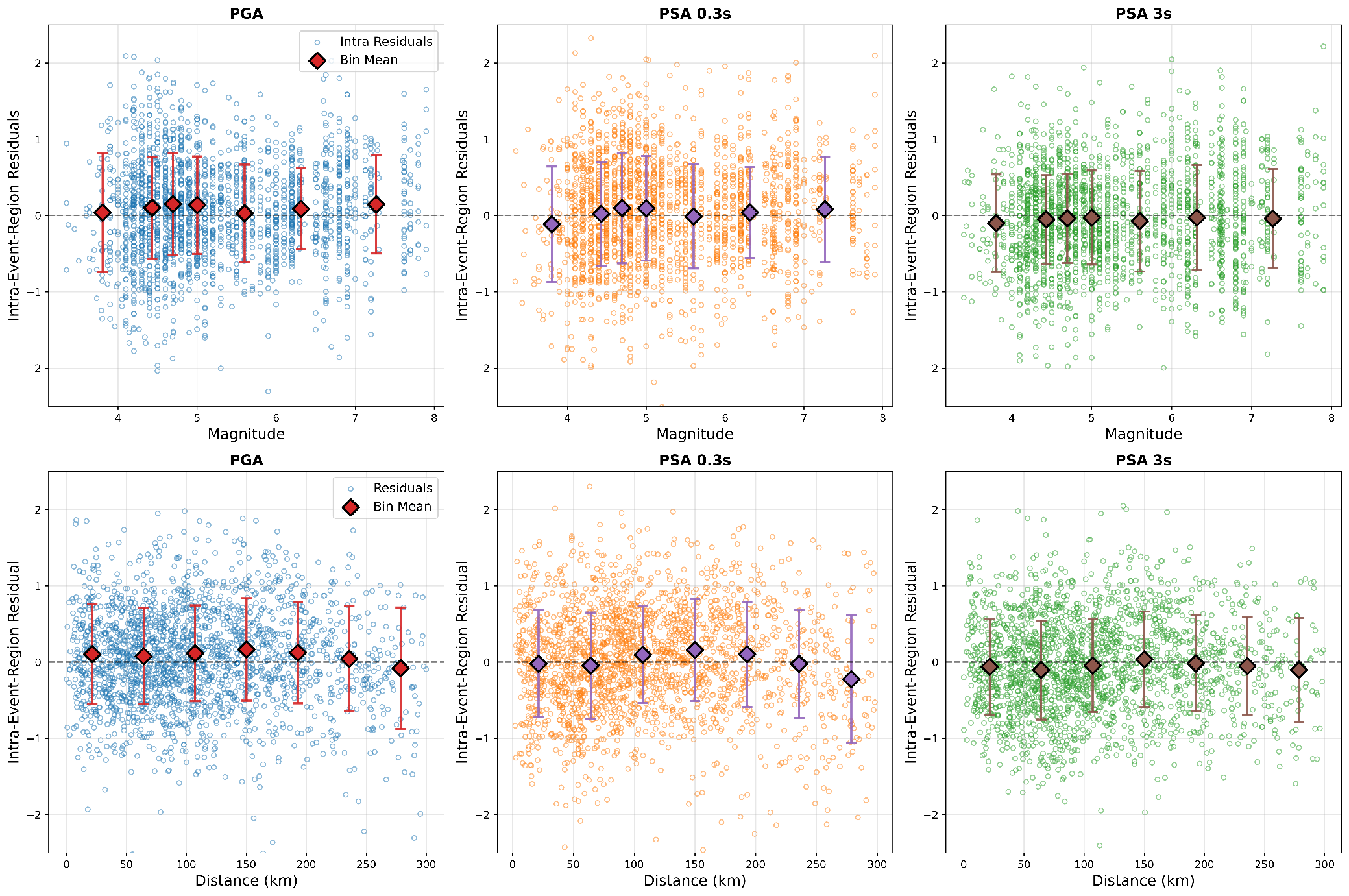}

Figure 9. Intra-region-event residuals plotted against magnitude (top
plots) and distance (bottom plots), respectively, for PGA, PSA at 0.3
and 3s.

\includegraphics[width=6.29792in,height=3.47014in]{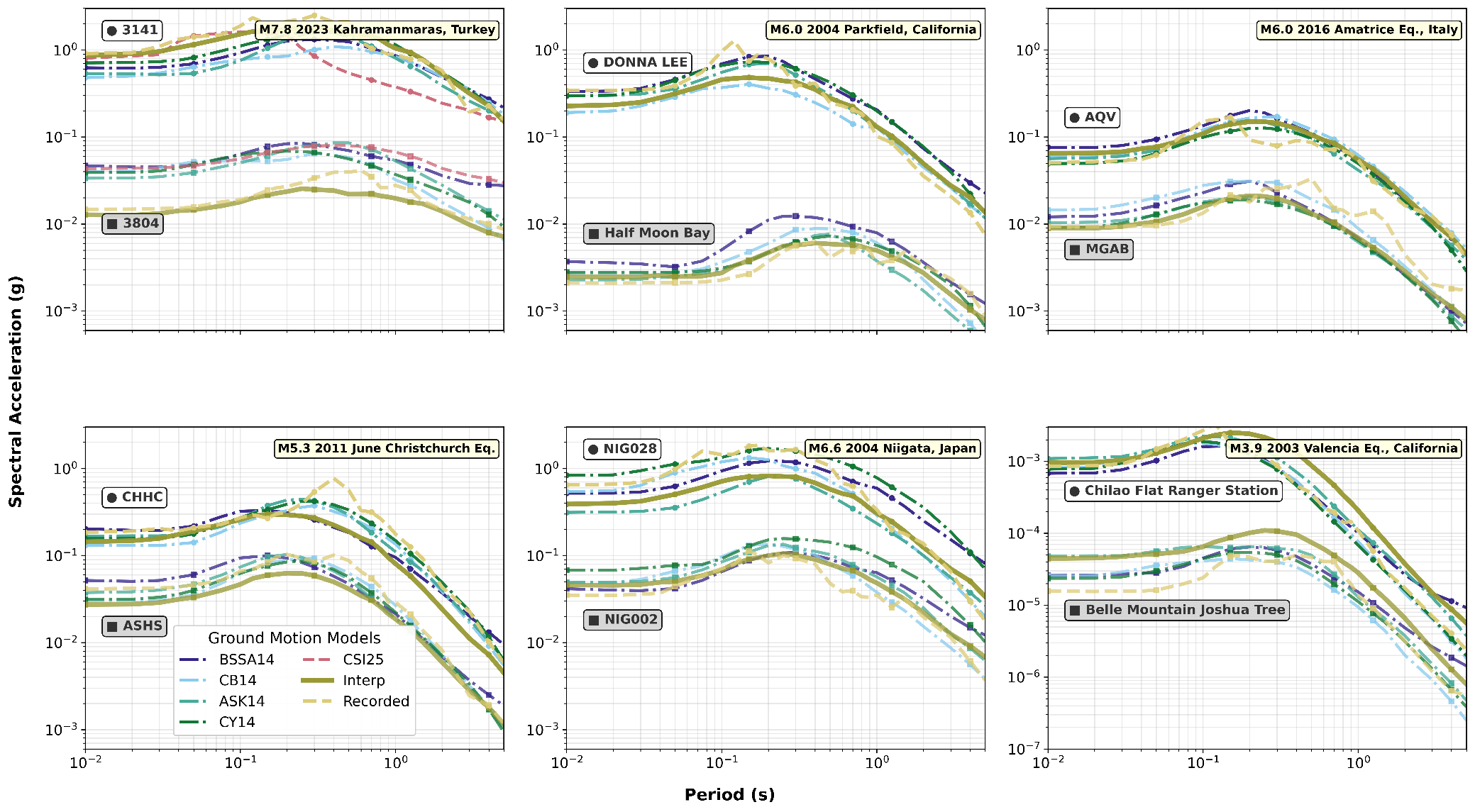}

Figure 10. Comparison of response spectra predictions from the
interpretable model and NGA-West2 GMMs (BSSA14, CB14, CY14, ASK14) and
Turkey model (CSI25) against recorded values for six earthquakes across
different regions.

\subsection{5.4 Comparison with other GMMs}

The interpretable model is evaluated against established NGA-West2 GMMs
across six diverse earthquake scenarios worldwide (Fig. 10).
Additionally, Turkey model (CSI25) is used in the comparison for the
Kahramanmaras earthquake. Each subplot presents spectral predictions at
two recording stations, with input parameters spanning magnitudes
3.9--7.8, rupture distances 0.1--248 km, all faulting mechanisms, depth
to top of rupture 0--13.4 km, shear-wave velocities 200--930 m/s, and
basin depths 10--1085 m. The comparative analysis demonstrates that the
interpretable model exhibits~strong agreement~with both recorded spectra
and NGA-West2 predictions across this comprehensive parameter space.
Similar robust performance is observed across the full range of spectral
periods and amplitude levels. The model shows consistent performance
across varying conditions, including small and large magnitudes, near-
and far-field scenarios, all styles of faulting (normal, reverse, and
strike-slip), surface to shallow rupture earthquakes, very soft soils to
hard rock sites, and shallow to deep basin depths. While acknowledging
the inherent~variability~in earthquake ground motion recordings, the
systematic agreement between the interpretable model predictions,
established GMMs, and observed data validates the
model\textquotesingle s capacity to capture spectral characteristics
across diverse seismological conditions. Furthermore, the smoothness
validation with BSSA14 demonstrated that the interpretable model is
significantly smooth, additional details are provided in the
supplementary.

Liu et al. (2025) achieves interpretability through equation sparsity,
but pathways can have implicit interactions within the discovered
equation. The developed model architecturally enforces independence via
separate pathways for each input using concurvity regularization. Future
symbolic learning work could integrate HazBinLoss-style weighting to
systematically prioritize sparse near-fault regions during equation
discovery.

\section{6. Conclusions}

This study develops an inherently interpretable and transparent
``glass-box'' ML-based GMM for 5\% damping to PGA, PGV, and PSA using
global active shallow crustal data, addressing the interpretability
limitations that hinder ML adoption in seismic risk assessment. We
demonstrate that ML models are not inherently "black-box"---instead,
interpretability depends on the architectural choices employed during
development.~Our approach achieves~interpretability by ensuring inputs
remain~disentangled~at every layer through
independent~additive~pathways. Additionally,~we introduce~a novel
HazBinLoss~function~to address~data~imbalance in seismic~datasets and
concurvity regularization to enforce orthogonality in pathway
decompositions.~This loss function combines~two components via a
hyperparameter ($\alpha$):

\begin{itemize}
\item
  Bin count component:~Applies~inverse weighting based on sample
  density~within~magnitude-distance bins.
\item
  Hazard-informed component:~Inspired by physics-constrained neural
  networks,~prioritizes~high-magnitude near-field records critical for
  hazard assessment.
\end{itemize}

Prediction results demonstrate that~the component scaling
of~our~developed model ($\alpha$ = 0.75)~exhibits~behavior consistent with
established seismological principles~and~period-dependent
characteristics. While MSE loss significantly underpredicts critical
scenarios (M\textsubscript{w} $\ge$ 7, R\textsubscript{rup} $\le$ 100km),
HazBinLoss~substantially~reduces this underprediction.~Notably,~the
interpretable model achieves performance comparable to established
NGA-West2 models and traditional feedforward networks with
HazBinLoss,~confirming~that interpretability through independent
additive pathways requires no accuracy sacrifice.

Ablation study simulates limited large-magnitude near-field data
scenarios by removing all records with M\textsubscript{w} $\ge$ 6 and
R\textsubscript{rup} $\le$ 100km. While MSE and $\alpha$ = 0~configurations~perform
poorly, other $\alpha$ values (0.25, 0.5, 0.75, 1.0)~maintain~significantly
better performance,~demonstrating~that HazBinLoss---particularly the
hazard-informed component---enables reliable model
performance~under~data-limited conditions. Nonlinear weighted
mixed-effects analysis reveals no bias or trends in inter-event or
intra-region-event~residuals,~confirming~unbiased
predictions.~Component-wise comparisons with an established NGA-West2
model demonstrated strong agreement across periods (R² \textgreater{}
0.8 in most cases), supporting the physical validity of the learned
pathway decomposition. Finally,~comprehensive validation shows~that
interpretable model predictions~remain~consistent with NGA-West2 GMMs
and observed data across diverse input conditions.

This work establishes a foundation for developing data-driven seismic
hazard assessment using interpretable ML architectures.~The pathway
orthogonality is achieved at the expense of higher MSE. It is
acknowledged that a strict interpretable architecture can't model
nonlinear pathway dependencies such as nonlinear site amplification.
Future research directions include relaxing strict interpretability
criteria and develop partially interpretable neural additive models,
incorporating Bayesian weights to quantify epistemic
uncertainty~for~individual pathway components, fine-tuning the model for
data-limited regions using HazBinLoss, coupling the regional flag with
the path terms, and conducting input ablation studies to~examine~how
component contributions vary across different~scenarios.~We envision
that this interpretable framework will enable broader ML adoption in
seismic risk assessment, providing both predictive accuracy and
transparency essential for engineering decision-making.

\textbf{Data Availability}: All the data used to support our results and
the final interpretable model are available at
\url{https://figshare.com/s/e2b052395beddf9ed6f0}.

\textbf{Author Contribution}

Vemula Sreenath: conceptualization, methodology, software, validation,
formal analysis, investigation, data curation, writing -- original
draft, visualization. Filippo Gatti: methodology, resources, writing,
review \& editing, project administration. Pierre Jehel: resources,
writing, review \& editing, project administration, funding acquisition.

\textbf{Declaration of generative AI in scientific writing}

While preparing the manuscript, the authors used generative AI tools
such as Claude 4.0 Sonnet to improve the readability and correct the
language. After using these tools, the authors reviewed and edited the
content as required and take full responsibility for the content of the
published article.

\textbf{Acknowledgments}

The authors appreciate the efforts by terms that processed and made
publicly available final usable databases: PEER NGA-West2, GeoNet New
Zealand, Turkey, and Italy. The revised manuscript is significantly
enhanced after incorporating comments from Prof. Fabrice Cotton and two
anonymous reviewers.

\textbf{Funding}: This research was carried out in the MINERVE project
number DOS0186108. The MINERVE project is supported by the French
government within the France 2030 framework: CORIFER AAP1-- Projet
«MINERVE». The authors thankfully acknowledge this support.

\textbf{Declaration of competing interest}: The authors declare that no
competing financial interests or personal relationships could have
influenced this work.

\textbf{References}

Abrahamson, N. A., Silva, W. J., \& Kamai, R. (2014). Summary of the
ASK14 ground motion relation for active crustal regions. Earthquake
Spectra, 30(3), 1025-1055. DOI: 10.1193/070913EQS198M.

Abrahamson, N. A., \& Youngs, R. R. (1992). A stable algorithm for
regression analyses using the random effects model. Bulletin of the
Seismological Society of America, 82(1), 505-510. DOI:
10.1785/BSSA0820010505.

Agarwal, R., Frosst, N., Zhang, X., Caruana, R., \& Hinton, G. E.
(2020). Neural additive models: interpretable machine learning with
neural nets. arXiv. arXiv preprint arXiv:2004.13912.

Akiba, T., Sano, S., Yanase, T., Ohta, T., \& Koyama, M. (2019, July).
Optuna: A next-generation hyperparameter optimization framework. In
Proceedings of the 25th ACM SIGKDD international conference on knowledge
discovery \& data mining (pp. 2623-2631). DOI:
10.48550/arXiv.1907.10902.

Al Atik, L., Abrahamson, N., Bommer, J. J., Scherbaum, F., Cotton, F.,
\& Kuehn, N. (2010). The variability of ground-motion prediction models
and its components. Seismological Research Letters, 81(5), 794-801. DOI:
10.1785/gssrl.81.5.794.

Alidadi, N., \& Pezeshk, S. (2025). State of the art: Application of
machine learning in ground motion modeling. Engineering Applications of
Artificial Intelligence, 149, 110534. DOI:
10.1016/j.engappai.2025.110534.

Ancheta, T.D., Darragh, R.B., Stewart, J.P., Seyhan, E., Silva, W.J.,
Chiou, B.S.J., Wooddell, K.E., Graves, R.W., Kottke, A.R., Boore, D.M.
and Kishida, T. (2014). NGA-West2 database. Earthquake Spectra, 30(3),
pp.989-1005. DOI: 10.1193/070913EQS197M.

Baker, J., Bradley, B., \& Stafford, P. (2021). Seismic hazard and risk
analysis. Cambridge University Press. DOI: 10.1017/9781108425056.

Bates, D., Mächler, M., Bolker, B., \& Walker, S. (2015). Fitting linear
mixed-effects models using lme4. Journal of Statistical Software, 67(1),
1--48. DOI: 10.18637/jss.v067.i01

Bindi, D. (2017). The predictive power of ground‐motion prediction
equations. Bulletin of the Seismological Society of America, 107(2),
1005-1011. DOI: 10.1785/0120160224.

Boore, D. M., Stewart, J. P., Seyhan, E., \& Atkinson, G. M. (2014).
NGA-West2 equations for predicting PGA, PGV, and 5\% damped PSA for
shallow crustal earthquakes. Earthquake Spectra, 30(3), 1057-1085. DOI:
10.1193/070113EQS184M.

Campbell, K. W., \& Bozorgnia, Y. (2014). NGA-West2 ground motion model
for the average horizontal components of PGA, PGV, and 5\% damped linear
acceleration response spectra. Earthquake Spectra, 30(3), 1087-1115.
DOI: 10.1193/062913EQS175M.

Chen, S., Liu, X., Fu, L., Wang, S., Zhang, B., \& Li, X. (2024).
Physics symbolic learner for discovering ground‐motion models via
NGA‐West2 database. Earthquake Engineering \& Structural Dynamics,
53(1), 138-151. DOI: 10.1002/eqe.4013.

Chiou, B. S. J., \& Youngs, R. R. (2014). Update of the Chiou and Youngs
NGA model for the average horizontal component of peak ground motion and
response spectra. Earthquake Spectra, 30(3), 1117-1153. DOI:
10.1193/072813EQS219M.

Derras, B., Bard, P. Y., \& Cotton, F. (2014). Towards fully data driven
ground-motion prediction models for Europe. Bulletin of Earthquake
Engineering, 12(1), 495-516. DOI: 10.1007/s10518-013-9481-0.

Derras, B., Bard, P. Y., \& Cotton, F. (2016). Site-condition proxies,
ground motion variability, and data-driven GMPEs: Insights from the
NGA-West2 and RESORCE data sets. Earthquake spectra, 32(4), 2027-2056.
DOI: 10.1193/060215EQS082M

Dhanya, J., \& Raghukanth, S. T. G. (2018). Ground motion prediction
model using artificial neural network. Pure and Applied Geophysics,
175(3), 1035-1064. DOI: 10.1007/s00024-017-1751-3.

Ding, J., Lu, D., \& Cao, Z. (2025). A hybrid nonparametric ground
motion model of power spectral density based on machine learning.
Computer‐Aided Civil and Infrastructure Engineering, 40(4), 483-502.
DOI: 10.1111/mice.13340.

Fayaz, J., Astroza, R., Angione, C., \& Medalla, M. (2024). Data-driven
analysis of crustal and subduction seismic environments using
interpretation of deep learning-based generalized ground motion models.
Expert Systems with Applications, 238, 121731. DOI:
10.1016/j.eswa.2023.121731.

Garson, G.D. (1991). Interpreting Neural Network Connection Weights. AI
Expert, 6, 47-51.

Gharagoz, M. M., Noureldin, M., \& Kim, J. (2025). Explainable machine
learning (XML) framework for seismic assessment of structures using
Extreme Gradient Boosting (XGBoost). Engineering Structures, 327,
119621. DOI: 10.1016/j.engstruct.2025.119621.

Harris, C.R., Millman, K.J., Van Der Walt, S.J., Gommers, R., Virtanen,
P., Cournapeau, D., Wieser, E., Taylor, J., Berg, S., Smith, N.J. and
Kern, R. (2020). Array programming with NumPy. Nature, 585(7825),
pp.357-362. DOI: 10.1038/s41586-020-2649-2.

Hastie, T., \& Tibshirani, R. (1986). Generalized additive models.
Statistical science, 1(3), 297-310. DOI: 10.1214/ss/1177013604.

Hunter, J. D. (2007). Matplotlib: A 2D graphics environment. Computing
in science \& engineering, 9(03), 90-95. DOI: 10.1109/MCSE.2007.55.

Hutchinson, J.A., Zhu, C., Bradley, B.A., Lee, R.L., Wotherspoon, L.M.,
Dupuis, M., Schill, C., Motha, J., Manea, E.F. and Kaiser, A.E., 2024.
The 2023 New Zealand ground‐motion database. Bulletin of the
Seismological Society of America, 114(1), pp.291-310. DOI:
10.1785/0120230184.

Idriss, I. M. (2014). An NGA-West2 empirical model for estimating the
horizontal spectral values generated by shallow crustal earthquakes.
Earthquake Spectra, 30(3), 1155-1177. DOI: 10.1193/1.2924362.

Kuehn, N. M., Bozorgnia, Y., Campbell, K. W., \& Gregor, N. (2023). A
regionalized partially nonergodic ground-motion model for subduction
earthquakes using the NGA-Sub database. Earthquake Spectra, 39(3),
1625-1657. DOI: 10.1177/87552930231180906.

Lanzano G., Ramadan F., Luzi L., Sgobba S., Felicetta C., Pacor F.,
D\textquotesingle Amico M., Puglia R., Russo E. (2022). Parametric table
of the ITA18 GMM for PGA, PGV and Spectral Acceleration ordinates.
Istituto Nazionale di Geofisica e Vulcanologia (INGV). DOI:
10.13127/ita18/sa\_flatfile.

Liu, X., Chen, S., Li, X., Fu, L., \& Cotton, F. (2025). A hybrid
symbolic learning approach for Ground-Motion model Development. Journal
of Asian Earth Sciences, 281, 106498.
https://doi.org/10.1016/j.jseaes.2025.106498.

Lundberg, S. M., \& Lee, S. I. (2017). A unified approach to
interpreting model predictions. Advances in neural information
processing systems, 30. DOI: 10.48550/arXiv.1705.07874.

McKinney, W. (2010). Data structures for statistical computing in
Python. Scipy, 445(1), 51-56. DOI: 10.25080/Majora-92bf1922-00a.

Meenakshi, Y., Vemula, S., Alne, A., \& Raghukanth, S. T. G. (2023).
Ground motion model for Peninsular India using an artificial neural
network. Earthquake Spectra, 39(1), 596-633. DOI:
10.1177/87552930221144330.

Miller, T. (2019). Explanation in artificial intelligence: Insights from
the social sciences. Artificial intelligence, 267, 1-38. DOI:
10.1016/j.artint.2018.07.007.

Mohammadi, A., Karimzadeh, S., Banimahd, S. A., Ozsarac, V., \&
Lourenço, P. B. (2023). The potential of region-specific
machine-learning-based ground motion models: application to Turkey. Soil
Dynamics and Earthquake Engineering, 172, 108008. DOI:
10.1016/j.soildyn.2023.108008.

Morikawa, N., \& Fujiwara, H. (2013). A new ground motion prediction
equation for Japan applicable up to M9 mega-earthquake. Journal of
Disaster Research, 8(5), 878-888.

Okazaki, T., Morikawa, N., Fujiwara, H., \& Ueda, N. (2021). Monotonic
neural network for ground‐motion predictions to avoid overfitting to
recorded sites. Seismological Society of America, 92(6), 3552-3564. DOI:
10.1785/0220210099.

Paszke, A., Gross, S., Massa, F., Lerer, A., Bradbury, J., Chanan, G.,
Killeen, T., Lin, Z., Gimelshein, N., Antiga, L. and Desmaison, A.
(2019). Pytorch: An imperative style, high-performance deep learning
library. Advances in neural information processing systems, 32. DOI:
10.48550/arXiv.1912.01703.

Ribeiro, M. T., Singh, S., \& Guestrin, C. (2016, August). "Why should I
trust you?" Explaining the predictions of any classifier. In Proceedings
of the 22nd ACM SIGKDD international conference on knowledge discovery
and data mining (pp. 1135-1144). DOI: 10.48550/arXiv.1602.04938.

Rudin, C. (2019). Stop explaining black box machine learning models for
high stakes decisions and use interpretable models instead. Nature
machine intelligence, 1(5), 206-215. DOI: 10.1038/s42256-019-0048-x.

Sandıkkaya, M. A., Güryuva, B., Kale, Ö., Okçu, O., İçen, A., Yenier,
E., \& Akkar, S. (2024). An updated strong-motion database of Türkiye
(SMD-TR). Earthquake Spectra, 40(1), 847-870. DOI:
10.1177/87552930231208158.

Sedaghati, F., \& Pezeshk, S. (2023). Machine learning--based ground
motion models for shallow crustal earthquakes in active tectonic
regions. Earthquake Spectra, 39(4), 2406-2435. DOI:
10.1177/87552930231191759.

Si, H., \& Midorikawa, S. (2000, January). New attenuation relations for
peak ground acceleration and velocity considering effects of fault type
and site condition. In Proceedings of 12th World Conference on
Earthquake Engineering (No. 0532). DOI: 10.3130/aijs.64.63\_2.

Siems, J., Ditschuneit, K., Ripken, W., Lindborg, A., Schambach, M.,
Otterbach, J., \& Genzel, M. (2023). Curve your enthusiasm: Concurvity
regularization in differentiable generalized additive models. Advances
in Neural Information Processing Systems, 36, 19029-19057.
arXiv:2305.11475.

Somala, S. N., Chanda, S., AlHamaydeh, M., \& Mangalathu, S. (2024).
Explainable XGBoost--SHAP machine-learning model for prediction of
ground motion duration in New Zealand. Natural Hazards Review, 25(2),
04024005. DOI: 10.1061/NHREFO.NHENG-1837.

Sreenath, V., Basu, J., \& Raghukanth, S. T. G. (2024). Ground motion
models for regions with limited data: Data‐driven approach. Earthquake
Engineering \& Structural Dynamics, 53(3), 1363-1375. DOI:
10.1002/eqe.4075.

United Nations Office for Disaster Risk Reduction. (2025). Global
Assessment Report on Disaster Risk Reduction 2025: Resilience pays:
Financing and investing for our future (Hazard explorations:
Earthquakes).
https://www.undrr.org/gar/gar2025/hazard-exploration/earthquakes. Last
accessed 20 June 2025.

Vemula, S., Yellapragada, M., Podili, B., Raghukanth, S. T. G., \&
Ponnalagu, A. (2021). Ground motion intensity measures for New Zealand.
Soil Dynamics and Earthquake Engineering, 150, 106928. DOI:
10.1016/j.soildyn.2021.106928.

Zhu, C., Cotton, F., Kawase, H., \& Nakano, K. (2023). How well can we
predict earthquake site response so far? Machine learning vs
physics-based modeling. Earthquake Spectra, 39(1), 478-504. DOI:
10.1177/87552930221116399.

\end{document}